\documentclass[10pt,twocolumn,letterpaper]{article}

\usepackage{wacv}
\usepackage{times}
\usepackage{epsfig}
\usepackage{graphicx}
\usepackage{amsmath}
\usepackage{amssymb}
\usepackage{subfigure}
\usepackage{caption}
\usepackage{colortbl}
\usepackage{enumitem}
\usepackage{relsize}
\usepackage[T1]{fontenc}
\usepackage[utf8]{inputenc}
\usepackage{color, colortbl}
\newtheorem{theorem}{Theorem}
\pdfoutput = 1

\wacvfinalcopy % *** Uncomment this line for the final submission

\def\wacvPaperID{516} % *** Enter the wacv Paper ID here

\ifwacvfinal\fi
\newcommand{\SKIP}[1]{} % Used to skip stuff we do not want type-set

\newcommand{\mbegin} {\left [ \begin{array}}

\newcommand{\mend}   {\end{array} \right ]}

\newcommand{\detbegin} {\left | \begin{array}}
\newcommand{\detend}   {\end{array} \right |}

\newcommand{\vbegin} {\left ( \begin{array}{c}}

\newcommand{\vend} {\end{array}\right )}

\def\squareforqed{\hbox{\rlap{$\sqcap$}$\sqcup$}}

\def\qed{\ifmmode\squareforqed\else{\unskip\nobreak\hfil
	\penalty50\hskip1em\null\nobreak\hfil\squareforqed
	\parfillskip=0pt\finalhyphendemerits=0\endgraf}\fi}

\newcommand{\showeqnlabel}{
	\hbox to 0pt{\quad\quad\relax\fbox{\scriptsize\rm\eqnlblx}%
	\gdef\eqnlblx{xxxx}}} \newcommand{\eqnlblx}{}

\def\@eqnnum{\rm (\theequation)\showeqnlabel}

\newcommand{\nofig}[1]{\centerline{\bf Figure here}}

\def\mat#1{\mathchoice{\mbox{\bf$\displaystyle\tt#1$}}
	{\mbox{\bf$\textstyle\tt#1$}}
	{\mbox{\bf$\scriptstyle\tt#1$}}
	{\mbox{\bf$\scriptscriptstyle\tt#1$}}}
\def\m#1{\protect\mat #1}

\begin{document}

%%%%%%%%% TITLE
\title{Non-Rigid Structure from Motion: Prior-Free Factorization Method Revisited}

\author{Suryansh Kumar\\
%Department of Information Technology and Electrical Engineering\\
Computer Vision Lab, ETH Z\"urich, Switzerland\\
{\tt\small sukumar@vision.ee.ethz.ch}
}

\maketitle
%\thispagestyle{empty}

%%%%%%%%% ABSTRACT
\begin{abstract}
   A simple prior free factorization algorithm \cite{dai2014simple} is quite often cited work in the field of Non-Rigid Structure from Motion (NRSfM). The benefit of this work lies in its simplicity of implementation, strong theoretical justification to the motion and structure estimation, and its invincible originality.  Despite this, the prevailing view is, that it performs exceedingly inferior to other methods on several benchmark datasets \cite{jensen2018benchmark,akhter2009nonrigid}. However, our subtle investigation provides some empirical statistics which made us think against such views. The statistical results we obtained supersedes Dai {\it{et al.}}\cite{dai2014simple} originally reported results on the benchmark datasets by a significant margin under some elementary changes in their core algorithmic idea \cite{dai2014simple}. Now, these results not only exposes some unrevealed areas for research in NRSfM but also give rise to new mathematical challenges for NRSfM researchers. We argue that by \textbf{properly} utilizing the well-established assumptions about a non-rigidly deforming shape i.e, it deforms smoothly over frames \cite{rabaud2008re} and it spans a low-rank space, the simple prior-free idea can provide results which is comparable to the best available algorithms. In this paper, we explore some of the hidden intricacies missed by Dai {\it{et. al.}} work \cite{dai2014simple} and how some elementary measures and modifications can enhance its performance, as high as approx. 18\% on the benchmark dataset. The improved performance is justified and empirically verified by extensive experiments on several datasets. We believe our work has both practical and theoretical importance for the development of better NRSfM algorithms. %Practically, it can also help improve the recently reported state-of-the-art \cite{kumar2017spatio, kumar2016multi, jensen2018benchmark} and other similar works in this field which are inspired by Dai et. al. work\cite{dai2014simple}.

%This paper revisits the well-known matrix factorization approach to Non-Rigid Structure-from-Motion (NRSfM) problem \cite{dai2014simple}. Our work can be conceived as an attempt to nullify the prevailing view about the ``prior-free" work \cite{dai2014simple} that it performs exceedingly inferior to other methods on several benchmark datasets \cite{jensen2018benchmark,akhter2009nonrigid}. At the same time, our work provides an in-depth understanding of ``prior-free'' method and how some powerful elementary measures and modifications can significantly enhance its performance, as high as 18\% on the popular datasets. We argue that by \textbf{properly} utilizing the well-established assumptions about a non-rigidly deforming shape i.e, it deforms smoothly over frames and it spans a low-rank space, the simple prior-free method can provide results which is comparable to the best available algorithms ---at the time of writing this paper. Similar to prior-free method we make ``low-rank'' shape assumption, and we show that a better solution to motion which satisfies our smooth motion assumption is already present within the estimated ``Gram matrix'', and an explicit motion regularization may not essentially be required. The improved performance is justified and empirically verified by extensive experiments on the publicly available benchmark datasets. Finally, our work conjecture some theoretical problems which we think needs attention for further developments in the matrix factorization approach to NRSfM. 
\end{abstract}

%%%%%%%%% BODY TEXT
\section{Introduction}
{\bf{Notation}}: \emph{The notation used in this paper is similar to Dai et al.  work \cite{dai2014simple} unless otherwise stated.} 

\begin{figure}
\centering
\subfigure[Paper Sequence]{\label{fig:fp1} {\includegraphics[width=0.24\textwidth, height=0.11\textheight]{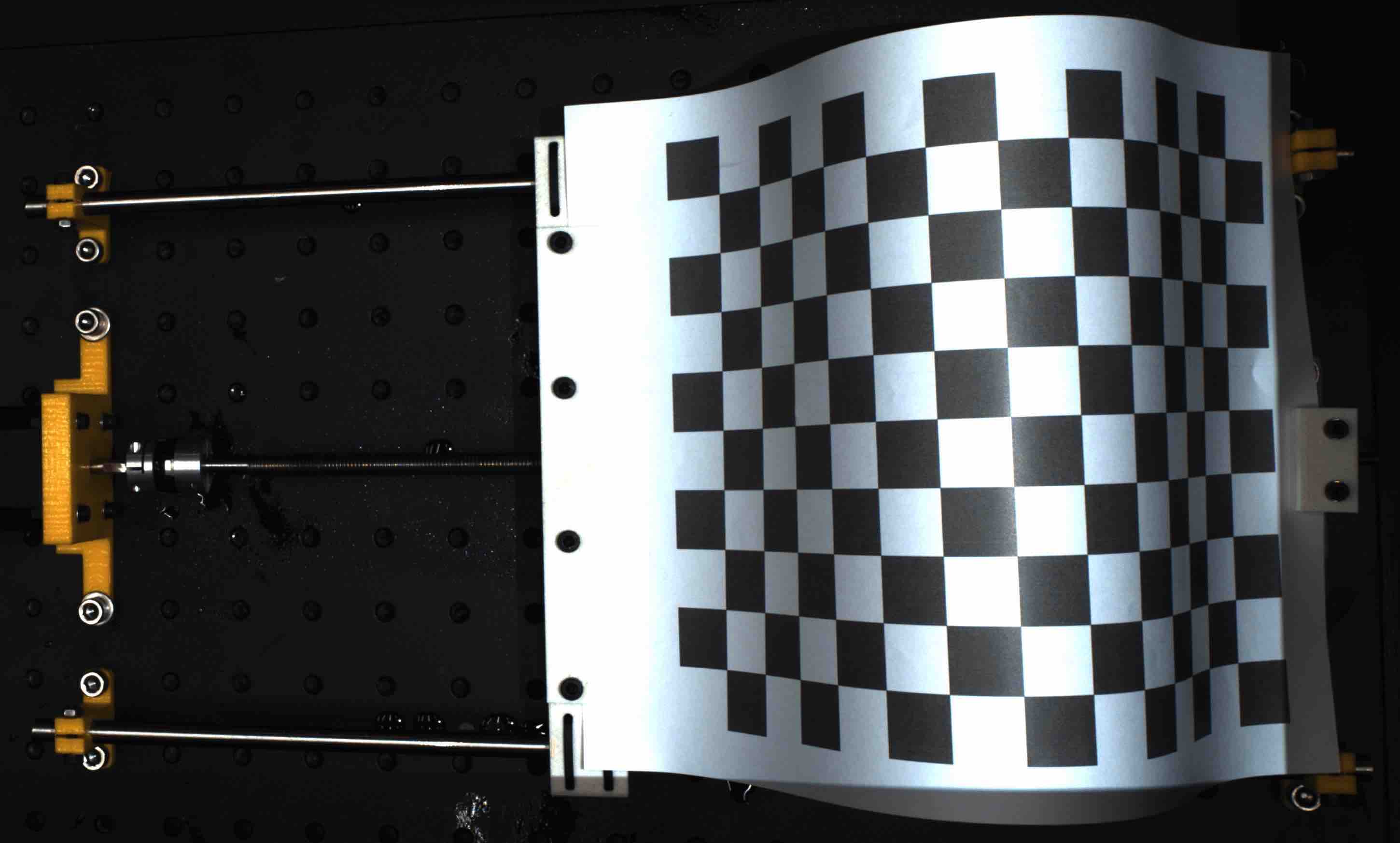}}}
\subfigure[3D Reconstruction]{\label{fig:fp2}{\includegraphics[width=0.16\textwidth, height=0.11\textheight]{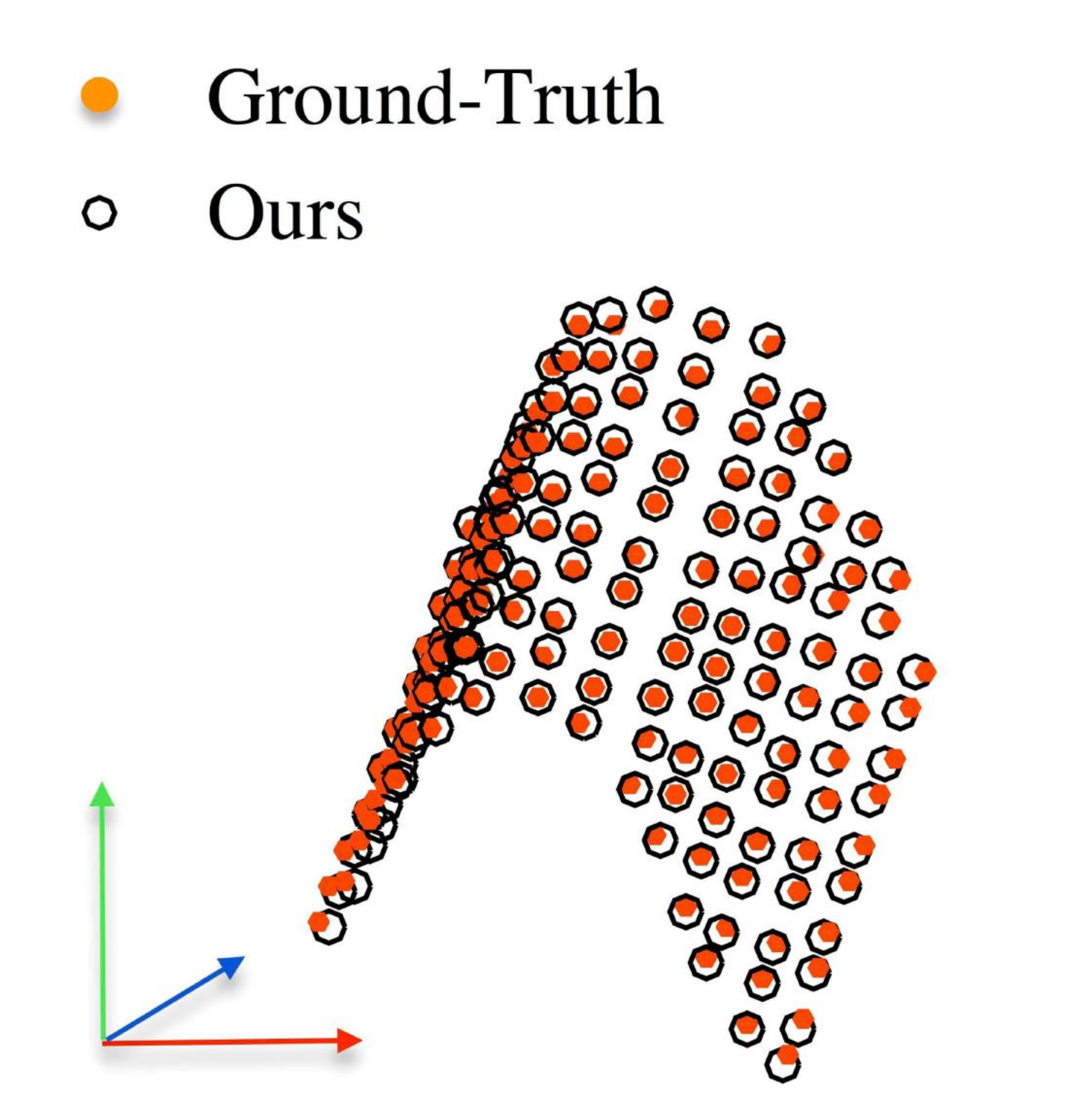}}}

\subfigure[Tearing Sequence] {\label{fig:fp3} {\includegraphics[width=0.24\textwidth, height=0.11\textheight]{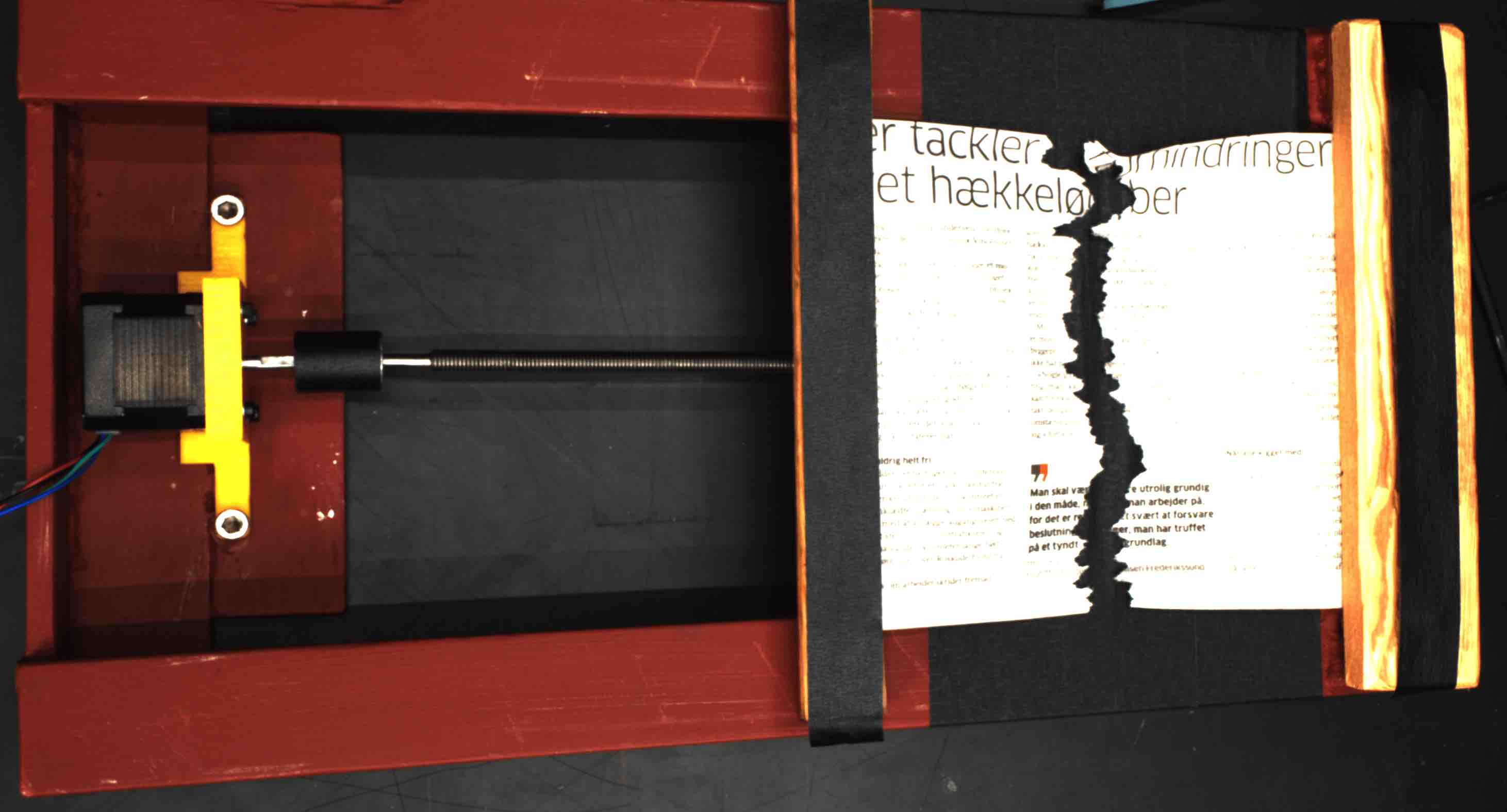}}}
\subfigure[3D Reconstruction] {\label{fig:fp4} {\includegraphics[width=0.17\textwidth, height=0.11\textheight]{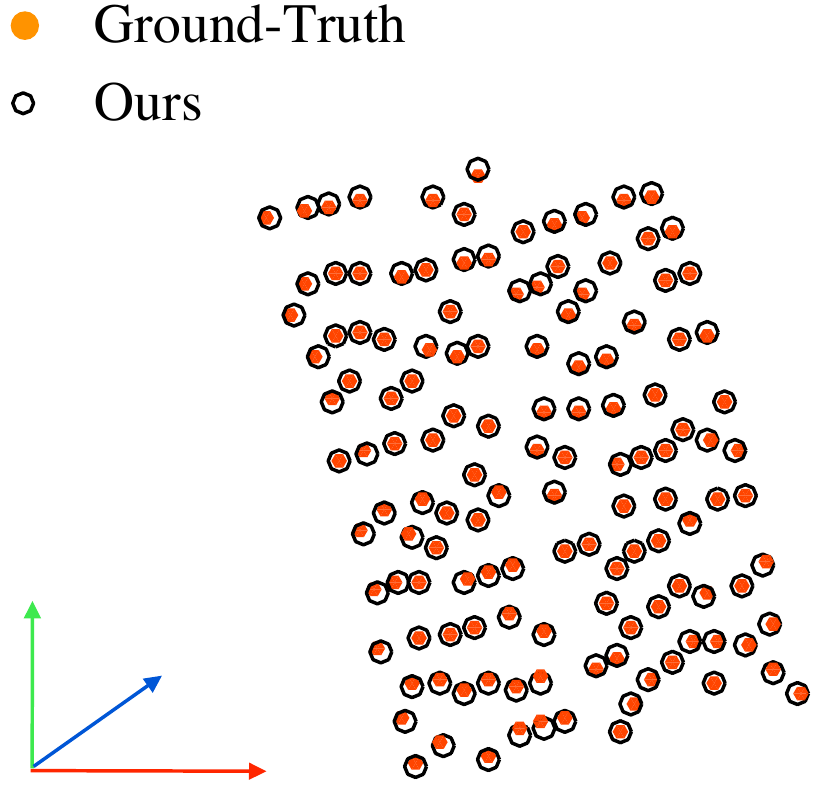}}}
\caption{\small{The method recovers 3D dimensional structure of the deforming object over multiple frames. Our elementary but powerful changes provides a substantial improvement in the  reconstruction accuracy than the previous results reported for ``prior-free" approach. The example images are taken from the recently released NRSfM Challenge Dataset \cite{jensen2018benchmark}. Our reconstruction results are nearly as good as the best performing algorithm without using very complex and involved mathematical optimization \cite{kumar2017spatio}.}}
\label{fig:firstpageresult}
\end{figure}

Non-rigid Structure from Motion (NRSfM) is a well-known problem in geometric computer vision \cite{bregler2000recovering, akhter2009nonrigid, dai2014simple, kumar2019superpixel, kumar2017monocular}. The goal of this problem is to reconstruct 3D structure of a deforming object using multiple frames. One of the most popular way to solve NRSfM is the matrix factorization approach. The matrix factorization approach to solve this problem dates back to 2000 \cite{bregler2000recovering} with no satisfactory solution in place until 2012. In the year 2012, Dai \etal  \cite{dai2012simple} proposed a ground-breaking approach to solve NRSfM. This method for solving NRSfM is now considered as a classical work in NRSfM \cite{dai2014simple}. In that paper, the camera motion is estimated by imposing the null space constraint and the rank-3 positive semi-definite matrix cone constraint on the Gram matrix ($\m Q_{\m k}$). Further, nuclear norm minimization of the reshuffled shape matrix ($\m S^{\sharp}$) was introduced to proffer stronger rank bound on the shape matrix for non-rigid shape estimation. The striking part of their work is that it not only challenged the myth of the inherent basis ambiguity in NRSfM \cite{xiao2004closed} but also supplied a practical ``prior-free" algorithm to solve NRSfM. Nonetheless, over years, it was observed that their remarkable theory performs poorly on benchmark datasets \cite{lee2013procrustean, jensen2018benchmark}. In this paper, our goal is to make ``prior-free idea'' work well on real world scenarios.

%without using complex mathematical notions such as subspace clustering \cite{kumar2017spatio}, manifold geometry \cite{rabaud2008re} \etc.

% \begin{figure}[t]
% \begin{center}
% \includegraphics[width=1.0\linewidth, height=0.22\textheight]{Figures/firstPagePlot.pdf}
% \caption{\small{The method recovers 3D dimensional structure of the deforming object over multiple frames. Our elementary rectification shows a substantial improvement in the camera motion and structure estimation than the previous reported results of ``prior-free" approach. The example images are taken from the recently released NRSfM Challenge Dataset \cite{jensen2018benchmark}. In the above results, filled and non-filled circles shows the ground-truth and reconstructed 3D points respectively. (a), (b) are the same results of the corresponding sequence from two different view-point.} \label{fig:firstpageresult}}
% \end{center}
% \end{figure}

Theoretically, the elementary idea of Dai \etal \cite{dai2014simple} conveniently encapsulates all the basic intuitions which are required to solve a general NRSfM problem. One may immediately argue on its performance when the deforming shape is composed of a union of low-rank subspace\cite{kumar2017spatio, kumar2016multi, zhu2014complex, kumar2018scalable, kumar2019jumping}. However, in this paper, we restrict our discussion to the classical representation of a NRSfM problem \cite{bregler2000recovering}, without paying much attention to, how clustering benefits 3D reconstruction of the non-rigid object and other such notions of compact data representation. The reason for this choice is that the improvement in the performance of a classical baseline shall benefit the methods built on top of it.

The main purpose of this work is to uncover some of the unexplored mathematical intricacies in the prior free factorization approach to NRSfM, and improve on the idea supplied by Dai \etal \cite{dai2014simple}. Our exposition leads to the possible reasons for its inferior performance on the benchmark datasets \cite{akhter2009nonrigid,jensen2018benchmark,torresani2008nonrigid}. It is shown in this paper that the rotation estimate using Dai \etal work \cite{dai2014simple} is \emph{not unique} under the same model complexity prior ($\m K\textrm{/rank}$), and they overlooked to utilize full correction matrix space \cite{brand2005direct}. Our investigation unveil the possibility of procuring motion that satisfies the well-known assumption of \emph{smooth} non-rigid deformation of the object \cite{rabaud2008re}. A simple search for the proper column-triplet (triads \cite{brand2005direct}) for the correction matrix ($\m G_{\m k}$) based on the smoothness of camera motion can indeed help improve the accuracy of the algorithm. Further, we argue that the weighted nuclear norm minimization of the shape matrix ($\m S^{\sharp}$) is a far better choice than its global trace norm minimization. Lastly, due to our extensive analysis, we are able to posit some unsolved issues in NRSfM under ``prior-free'' idea which needs attention for further progress in this field.

In this paper, it is not claimed that we achieve state-of-the-art results on the benchmark datasets using our new approach. However, we empirically show that we can get very close to the best performing approaches and the difference is not very great, without the employment of complex and involved mathematical optimization \cite{kumar2017spatio,lee2013procrustean}. In this paper, we also argue that the inferior performance of ``prior-free" method may not be due the proposed theoretical idea but because they overlooked some of the mathematical construction in their own formulation, and missed on properly utilizing the well-known assumptions about non-rigidly moving object \ie, \emph{smooth deformation} \cite{rabaud2008re} and \emph{low-rank shape} \cite{dai2014simple}. Hence, the conclusion, understanding, and use of simple ``prior-free'' algorithm to NRSfM is not complete and precise. Through this work, we try to amend and nullify the prevailing perception about the ``prior-free" approach, and how it can be used to its maximum potential. We feel that our paper touches some critical points which are essential to establish a theoretical closure to some of the elementary problems within the factorization approach to NRSfM.

\noindent
{\bf{Contribution:}} Firstly, our work postulates some rectification to the usage of ``Intersection Method" \cite{dai2014simple} to compute camera motion. With the suitable example, we establish that the generalization made on the rotation matrix estimation by Dai \etal work  \cite{dai2014simple}  is \emph{not convincing} and therefore, the knowledge about the strength of ``Intersection theorem'' is not completely exploited. Secondly, we provide an analytic solution to estimate suitable rotation using Intersection theorem and \textbf{conjecture} some challenges associated with it. Lastly, we propose a weighted nuclear norm minimization problem to estimate non-rigid 3D shape. Our approach shows a substantial improvement in the 3D reconstruction accuracy ({\bf{nearly 18\%}}). Moreover, we observed performance improvement in the case of noisy and missing trajectories \S \ref{ss:missing} (under minor adjustment) using our method.
%We also observed improvements in the performance of the algorithm in the presence of noisy data \S \ref{ss:noisy} and missing data \S \ref{ss:missing} (with a minor adjustment).

In this work, our attempt is to make the baseline method\footnote{By baseline, we mean the methods that solve NRSfM using its classical representation  $\m W = \m R \m S$ that have withstood the test of time \cite{tomasi1992shape,bregler2000recovering}.} more accurate, both in terms of understanding and performance, subject to the mathematical simplicity. To achieve this, we attempt to avoid the usage of complex mathematical notions such as union of independent subspace, dependent subspace representation \cite{zhu2014complex,kumar2017spatio,larsson2017compact}, procrustean normal distribution  \cite{lee2013procrustean}, kernelization \cite{gotardo2011kernel} \etc. Hence, it is simple to understand the theoretical and practical justification of our method. We show that by applying simple but powerful logical and mathematical modifications to the prior free idea \cite{dai2014simple},  we can get close to or even perform better at times than the best available algorithms on the benchmark datasets. 
%Additionally, our approach shall help improve the other state-of-the-art methods built on top of the targeted baseline \cite{dai2014simple}.

% We show improvement both in the rotation estimation and the 3D reconstruction accuracy using our approach under the same model complexity value (\ie $\m K$) used in the original Dai \etal work \cite{dai2014simple}. This help establish that under different choice of column triplet for $\m G_{\m k}$ the rotation estimates are different and therefore, \emph{the generalization made on the uniqueness of rotation estimate by Dai et al. is not proper and therefore, the arc of knowledge coming from $\m Q_{\m k}$ is not completely revealed}.

% and then proposes a better way to solve for 3D structure estimation. Lastly, this paper concludes with some open problems
% and through this work we are tying to reveal some of them for better interpretation. 

\section{Representation and Motion Estimation}

\paragraph{1. Classical Representation:}
Tomasi and Kanade factorization method to structure-from-motion under orthographic camera projection appropriately summarizes the behavior of the 3D points over frames \cite{tomasi1992shape}. The relation between 3D shape, motion and its projection over frames was defined as
\begin{equation}
\m W = \m R \m S
\end{equation}
where, $\m W \in \mathbb{R}^{2\m F \times \m P}$ is the measurement matrix formed by stacking all the image coordinates ($\mathbf{x} = [\m u, \m v]^{\m T}$) for `$\m P$' points along `$\m F$' rows \ie, total number of frames. $\m R = \textrm{{\fontfamily{cmtt}\selectfont blockdiagonal}}(\m R_1, \m R_2,..,\m R_{\m F}) \in \mathbb{R}^{2 \m F \times 3 \m F}$ denotes the orthographic camera rotation matrix with each $\m R_{\m i} \in \mathbb{R}^{2 \times 3}$ as per frame rotation. $\m S \in \mathbb{R}^{3 \m F \times \m P}$ represent the shape matrix with each row triplet as a 3D shape.  This representation was later extended by Bregler \etal \cite{bregler2000recovering} to recover non-rigid 3D shapes. More concretely,
\begin{equation}\label{eq:2}
\begin{aligned}
& \displaystyle \m W =\begin{bmatrix}
    \mathbf{x}_{11} \dots \mathbf{x}_{1 \m P}\\
    \dots\\
   \mathbf{x}_{\m F 1} \dots \mathbf{x}_{\m F \m P}
\end{bmatrix}=\begin{bmatrix} \m R_1 \m S_1 \\..\\ \m R_{\m F} \m S_{\m F} \end{bmatrix}=\begin{bmatrix}
    \mathbf{c}_{11} \m R_1 \dots \mathbf{c}_{1 \m K} \m R_1\\
    \dots\\
   \mathbf{c}_{\m F 1} \m R_{\m F} \dots \mathbf{c}_{\m F \m K} \m R_{\m F}
\end{bmatrix}\begin{bmatrix} \m B_1 \\ .. \\ \m B_{\m K} \end{bmatrix} \\
& \displaystyle \Rightarrow \m W = \m R( \m C \otimes \m I_{3}) \m B = \Pi \m B
\end{aligned}
\end{equation}
The matrix `$\m B$' and `$\m C$' are composed of shape bases and shape coefficients respectively, with `$\m K$' as the number of shape bases. `$\otimes$' denotes the kronecker product and `$\m I_{3}$' is a $3 \times 3$ identity matrix. It is evident from the above formulation that the rank of $\m W \leq 3 \m K$ and also $\textrm{rank}(\m S) \leq 3 \m K$. However, $\m S$ is not a general rank $3 \m K$ matrix but own a special structure due to $\m C \otimes \m I_{3}$ factor \cite{dai2014simple}.
\paragraph{2. Null Space Representation of the Orthonormality Constraint:}\label{ss:rrot}
An initial step in the factorization approach to NRSfM is to perform a  rank $3 \m K$ decomposition of the measurement matrix $\m W$ via singular value decomposition ({\fontfamily{cmtt}\selectfont svd}) \ie $\m W = \hat{\Pi}\hat{ \m B}$.
This is then followed by the estimation of Euclidean corrective matrix `$\m G$' to solve rotation and 3D structure. The main reason for such a procedure is due to the fact that the singular value decomposition of `$\m W$' matrix is not unique as any non-singular matrix $\m G \in \mathbb{R}^{3\m K \times 3 \m K}$ in between the two matrices $\hat{\Pi}$ and $\hat{ \m B}$  can form a valid factorization. Mathematically,
\begin{equation}\label{eq:3}
\m W \equiv \hat{\Pi}\hat{ \m B} = (\hat{\Pi}\m G) (\m G^{-1} \hat{ \m B}) = {\Pi}{\m B}
\end{equation}
Now, once we are able to solve $\m G$ correctly, then rotation and shape can be estimated using the above relations \cite{bregler2000recovering}. To solve $\m G$, orthonormality constraints are imposed \ie $\m R_{\m i} \m R_{\m i}^{\m T} = \m I_2$. Representing the $\m i^{\textrm{th}}$ double row of $\hat{\Pi}$ as $\hat{\Pi}_{2\m i -1: 2\m i} \in \mathbb{R}^{2 \times 3\m K}$ and $\m G_{\m k} \in \mathbb{R}^{3 \m K \times 3}$ as the $\m k^{\textrm{th}}$ column triplet of $\m G$, then using Eq:(\ref{eq:2}) and Eq:(\ref{eq:3}) we can write
\begin{equation}\label{eq:4}
\hat{\Pi}_{2\m i -1: 2\m i}\m G_{\m k} = \m c_\textrm{ik} \m R_{\m i},  \forall ~\m i = \{1, 2, .., \m F \}, \m k = \{1, 2, .., \m K\}
\end{equation}
Multiplying both sides by $\m R_{\m i}^{\m T}$ from right side gives
\begin{equation}\label{eq:5}
\begin{aligned}
& \displaystyle \hat{\Pi}_{2\m i -1: 2\m i}\m G_{\m k} \m G_{\m k}^{\m T}\hat{\Pi}_{2\m i -1: 2\m i}^{\m T} = \m c_\textrm{ik}^{2} \m I_{2} \\
& \displaystyle \text{This leads to two linear equation constraint }\\
& \displaystyle \hat{\Pi}_{2\m i -1} \m Q_{\m k} \hat{\Pi}_{2\m i -1}^{\m T} = \hat{\Pi}_{2\m i} \m Q_{\m k} \hat{\Pi}_{2\m i}^{\m T}, 
~~\hat{\Pi}_{2\m i -1} \m Q_{\m k} \hat{\Pi}_{2\m i}^{\m T} = 0
\end{aligned}
\end{equation}
where, $\m Q_{\m k} \in \mathbb{R}^{3 \m K \times 3 \m K} = \m G_{\m k}\m G_{\m k}^{\m T}$. Using the algebraic relation $\textrm{vec}(\m A \m X \m B^{\m T})$ = $(\m B \otimes \m A)\textrm{vec}(\m X)$, Dai \etal transformed these constraints (Eq:\ref{eq:5}) to a null space representation as follows:
\begin{equation}
\begin{aligned}
& \displaystyle\begin{bmatrix}
     \hat{\Pi}_{2\m i -1} \otimes \hat{\Pi}_{2\m i -1} - \hat{\Pi}_{2\m i} \otimes \hat{\Pi}_{2\m i}\\
     \hat{\Pi}_{2\m i -1} \otimes \hat{\Pi}_{2\m i} 
\end{bmatrix}\textrm{vec}(\m Q_{\m k}) = \m A \textrm{vec}(\m Q_{\m k}) = 0 
\end{aligned}
\end{equation}

Using the above form and previous work in NRSfM \cite{xiao2004closed}, Dai \etal proposed the \emph{intersection theorem} and supplied a SDP solution to estimate the $\m Q_{\m k}$ matrix and the Euclidean corrective matrix $\m G_{\m k}$ using {\fontfamily{cmtt}\selectfont svd()}.
\begin{theorem}\label{th:1}
Intersection Theorem: Under non-generate and noise-free conditions, any correct solution of $\m Q_{\m k}$ must lie in the intersection of the $(2\m K^{2}- \m K)$ dimensional null-space of $\m A$ and a rank 3 positive semi-definite matrix cone \ie $\m Q_{\m k}$ must belong to
\end{theorem}
\begin{equation}\label{eq:7}
\{\m A \textrm{vec}(\m Q_{\m k})\} \cap \{ \m Q_{\m k} \succeq 0 \} \cap \{ \textrm{rank}( \m Q_{\m k}) = 3\}
\end{equation}

\begin{figure}
\centering
 \subfigure [\label{fig:a1}] {\includegraphics[width=0.30\textwidth, height=0.12\textheight]{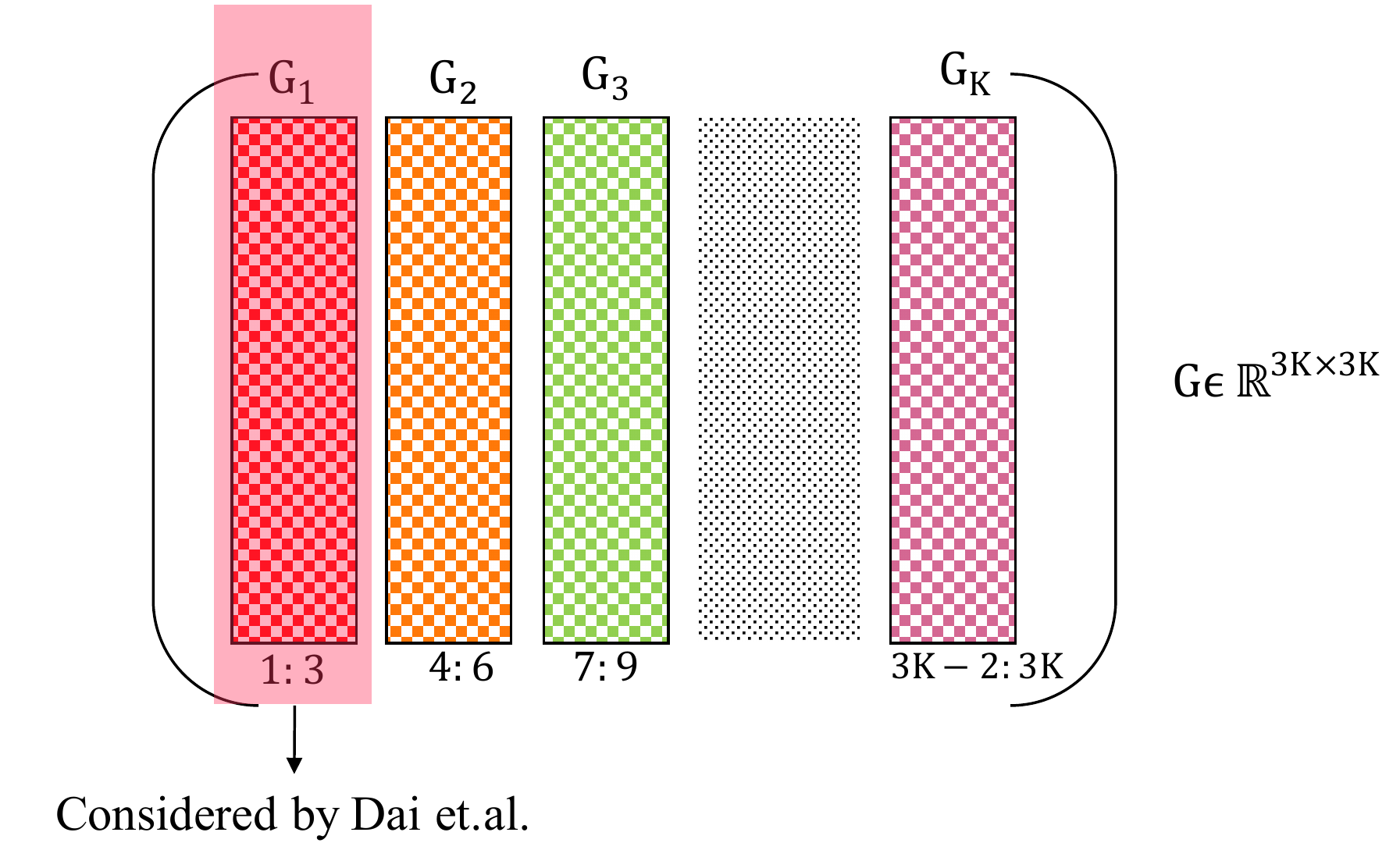}}
 \caption{ \small{(a) The column triplet (1:3) of euclidean corrective matrix ($\m G_{\m k}$) used by Dai \etal work \cite{dai2014simple} shown in red shade. It is stated with the notion that there is no loss of generality to choose $\m G_{1}$. However, choosing other column triplet may result in better rotation and shape estimate as shown in Figure \ref{fig:a2} and \ref{fig:a3}}}
\end{figure}

\noindent
{\bf{Dai \etal solution to rotation}}: They proposed that once the $\m Q_{\m k}$ is solved, rather than solving for full Euclidean corrective matrix $\m G \in \mathbb{R}^{3 \m K \times 3 \m K}$, use {\fontfamily{cmtt}\selectfont svd()} to extract rank 3 $\m G_{\m k}$. The solved $\m G_{\m k} \in \mathbb{R}^{3 \m K \times 3}$ can then be use to find $\m R$ (Eq:\ref{eq:4}) up to sign ($\m c_\textrm{ik}$). The method quote ``we adopt a simpler approach that directly computes the camera motion $\m R$ from single column-triplet $\m G_{\m k}$ without need to fill in a big and full $\m G$ matrix''. Naturally, this single column-triplet is chosen to be the first column-triplet $(\m G_{1})$ of the $ \m G$ matrix (see Fig:\ref{fig:a1}). Now, such strategy give rise to few legitimate {\emph{concerns}}
\begin{enumerate}[nolistsep, itemindent=0em, label=(\alph*)]
\item When each column triplet $\{\m G_{\m i}\}_{\m i=1}^{\m K}$ qualifies for a suitable correction matrix, then why $\m G_{1}$ has a high preference? Are we loosing useful information by such unwarranted preference? 
\item Will each $\{\m G_{\m i}\}_{\m i=1}^{\m K}$ provide the same solution to the rotation matrix?
\item Generally, most real world deformations are smooth in nature \cite{rabaud2008re}. Whether such solution to rotation is good enough for the smooth deformation assumption?
\end{enumerate}
Dai \etal overlooked all these intrinsic issues to solve rotation using their proposed intersection theorem.

\begin{figure}
\centering
\subfigure[$\m R \gets \m G_{1}$]{\label{fig:rot1} {\includegraphics[width=0.09\textwidth, height=0.09\textheight]{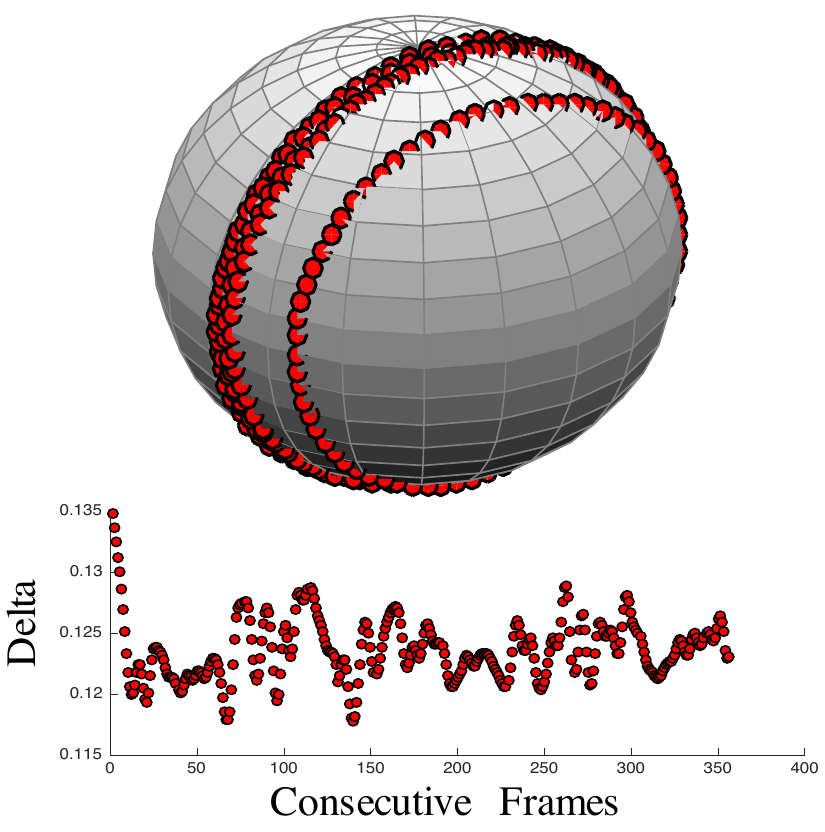}}}
\subfigure[$\m R \gets \m G_{2}$]{\label{fig:rot2}{\includegraphics[width=0.09\textwidth, height=0.09\textheight]{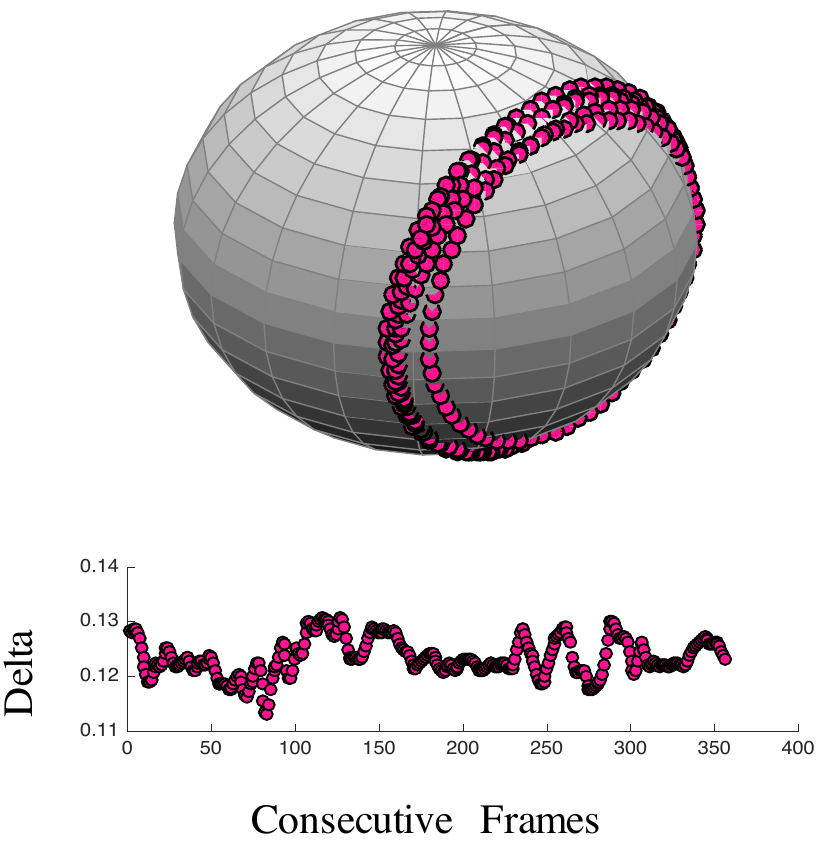}}}
\subfigure[$\m R \gets \m G_{3}$]{\label{fig:rot3} {\includegraphics[width=0.09\textwidth, height=0.09\textheight]{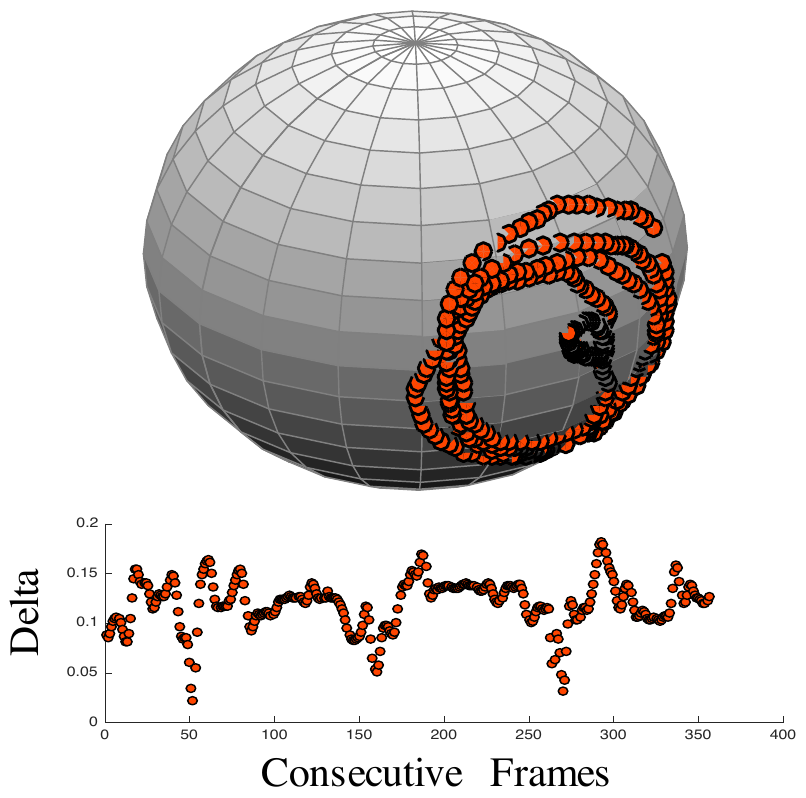}}}
\subfigure[$\m R \gets \m G_{4}$] {\label{fig:rot4}{\includegraphics[width=0.09\textwidth, height=0.09\textheight]{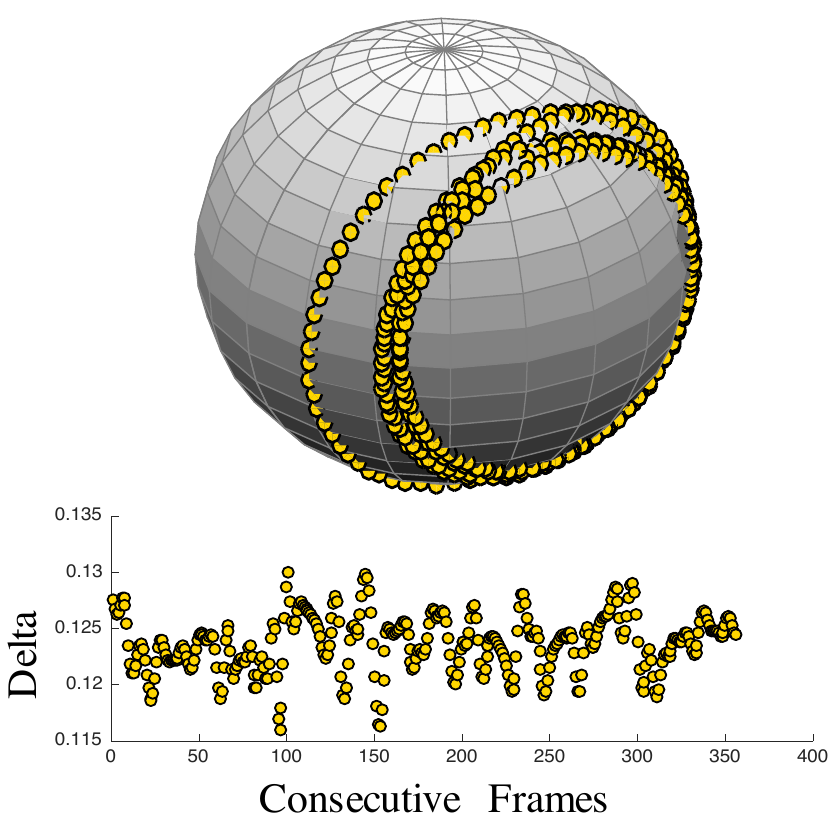}}}

\subfigure[$\m R \gets \m G_5$] {\label{fig:rot5} {\includegraphics[width=0.09\textwidth, height=0.09\textheight]{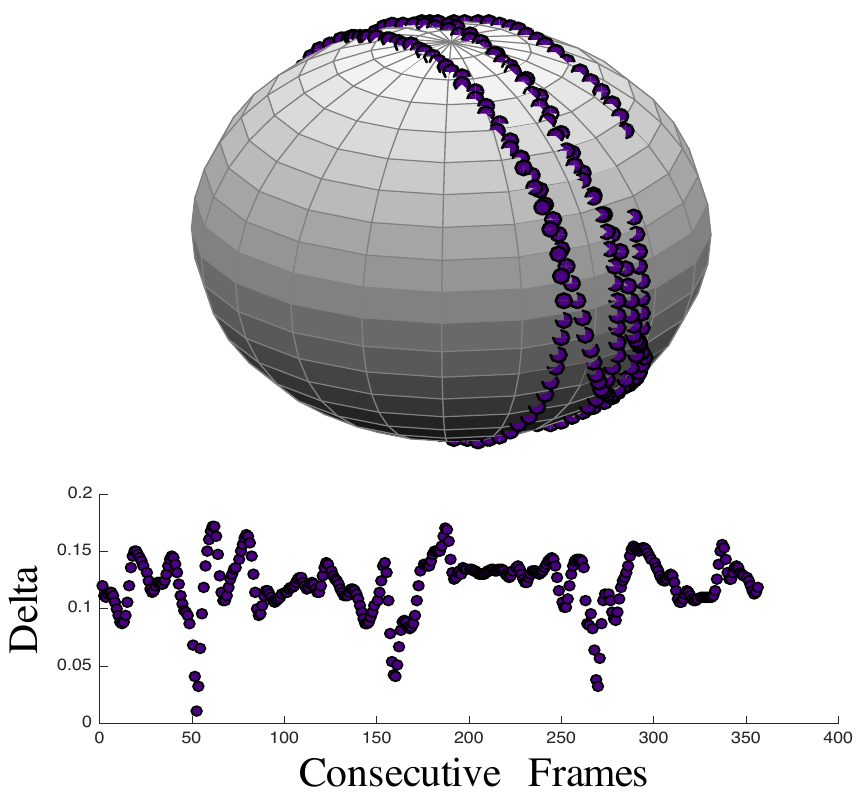}}}
\subfigure[$\m R \gets \m G_6$] {\label{fig:rot6} {\includegraphics[width=0.09\textwidth, height=0.09\textheight]{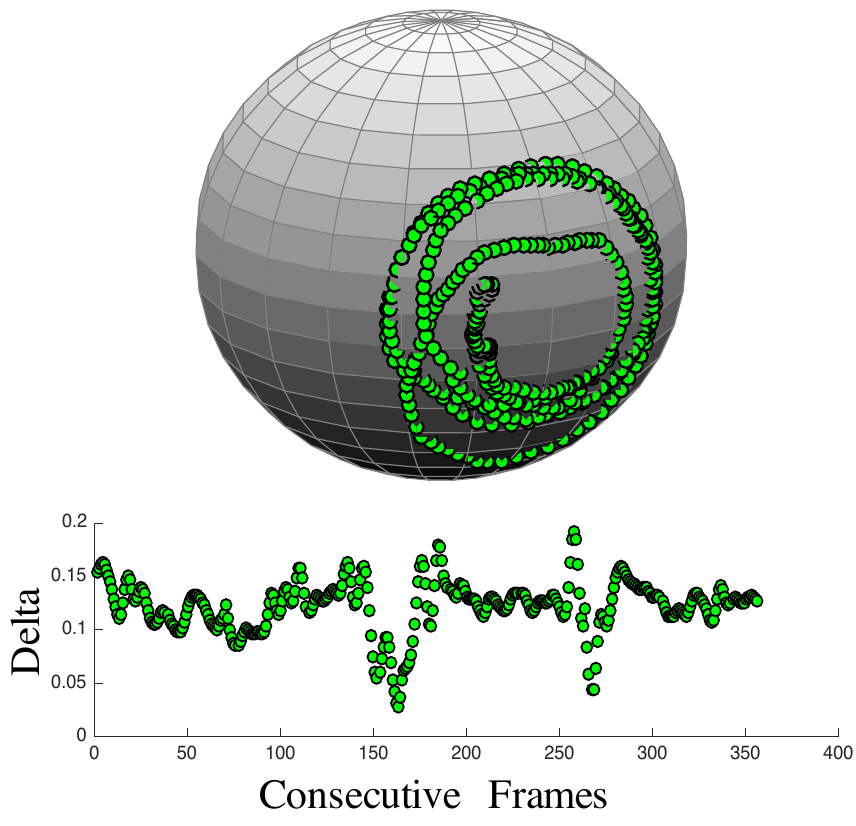}}}
\subfigure[$\m R \gets \m G_7$] {\label{fig:rot7}{\includegraphics[width=0.09\textwidth, height=0.09\textheight]{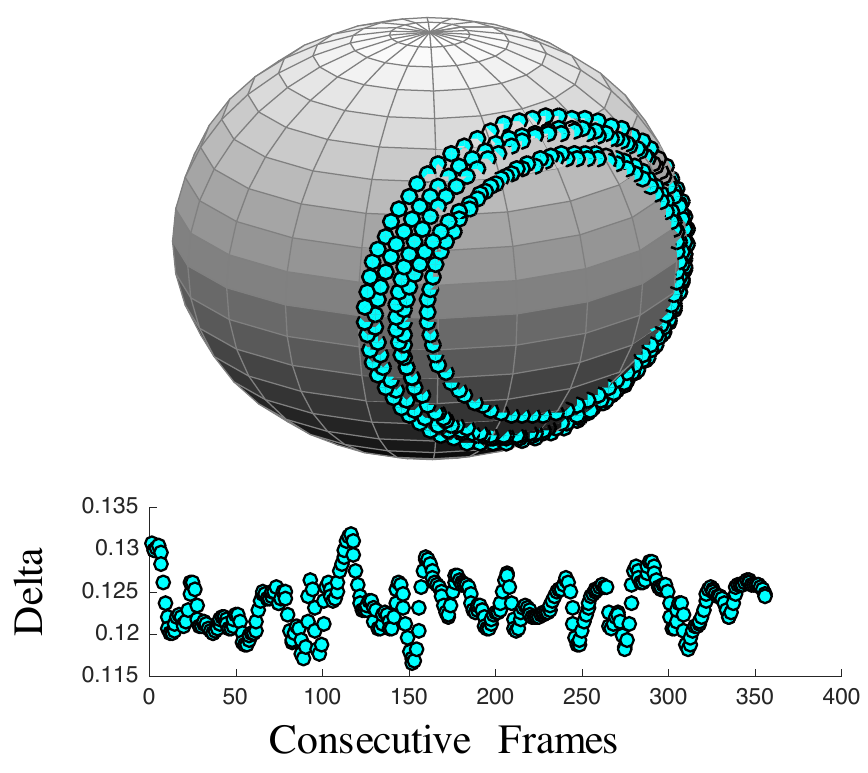}}}
\subfigure [$\m R \gets \m G_8$] {\label{fig:rot8} {\includegraphics[width=0.09\textwidth, height=0.09\textheight]{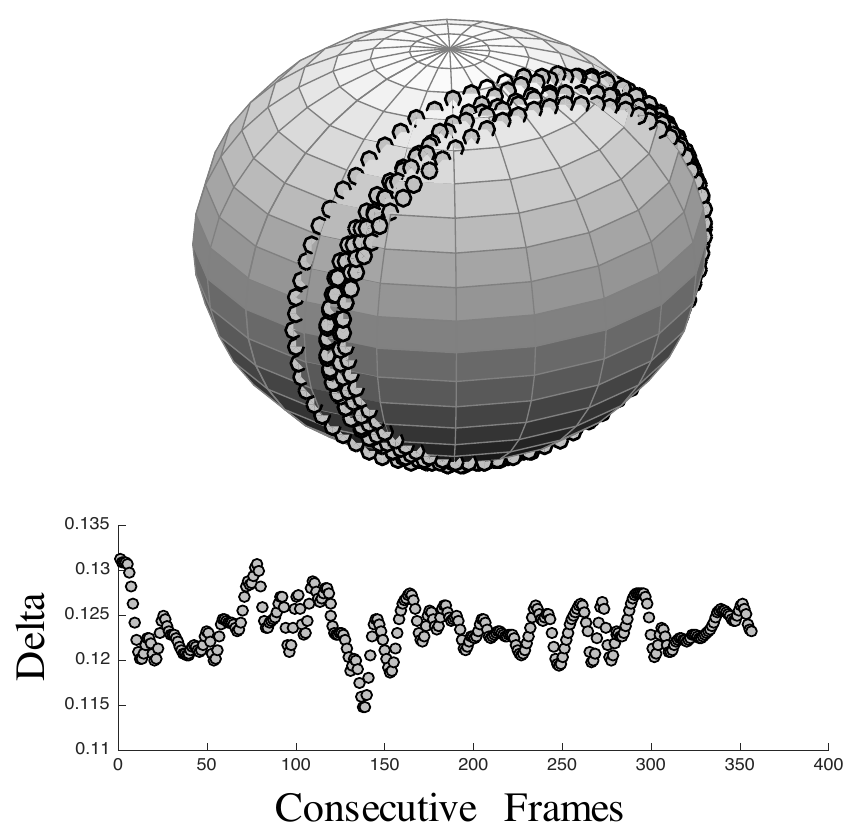}}}

\subfigure[$\m R \gets \m G_9$] {\label{fig:rot9} {\includegraphics[width=0.09\textwidth, height=0.09\textheight]{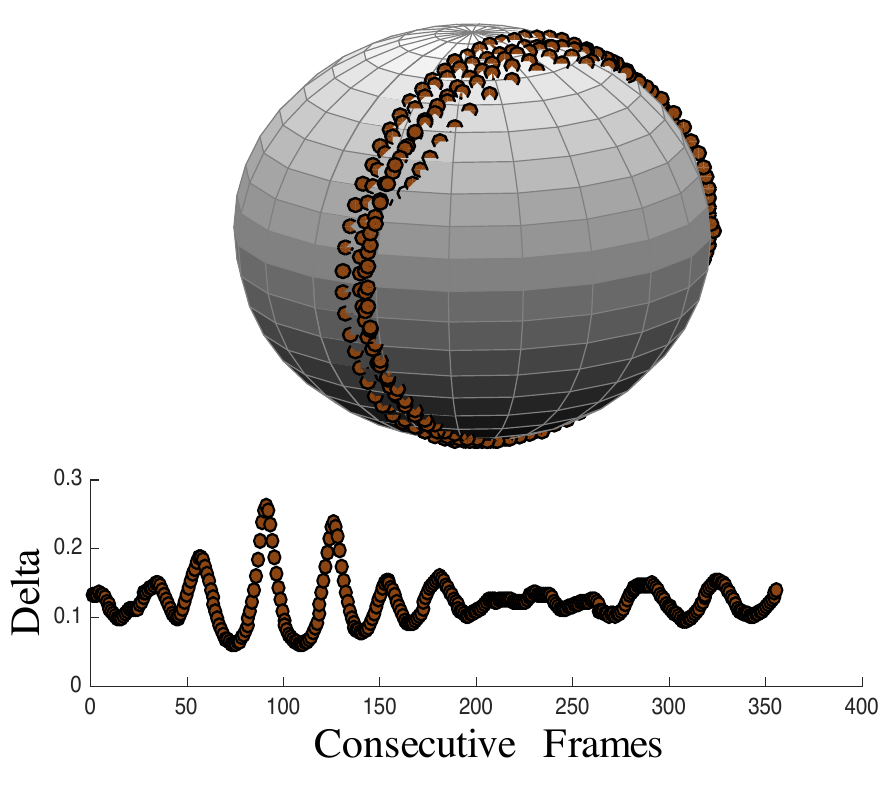}}}
\subfigure[$\m R \gets \m G_{10}$] {\label{fig:rot10} {\includegraphics[width=0.09\textwidth, height=0.09\textheight]{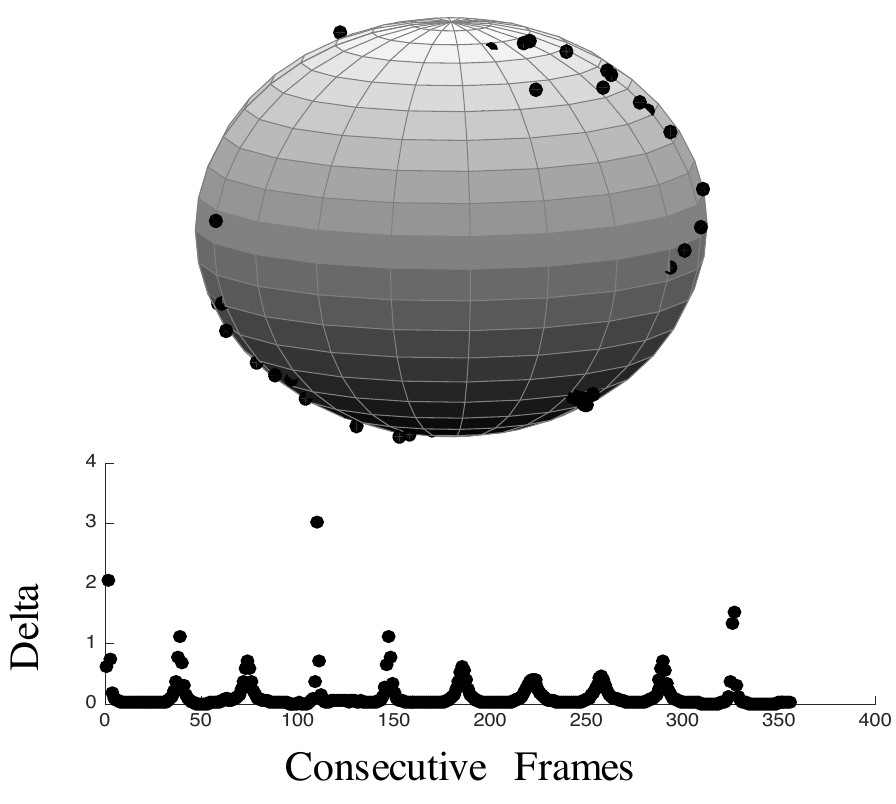}}}
\subfigure[$\m R \gets \m G_{11}$] {\label{fig:rot11}{\includegraphics[width=0.09\textwidth, height=0.09\textheight]{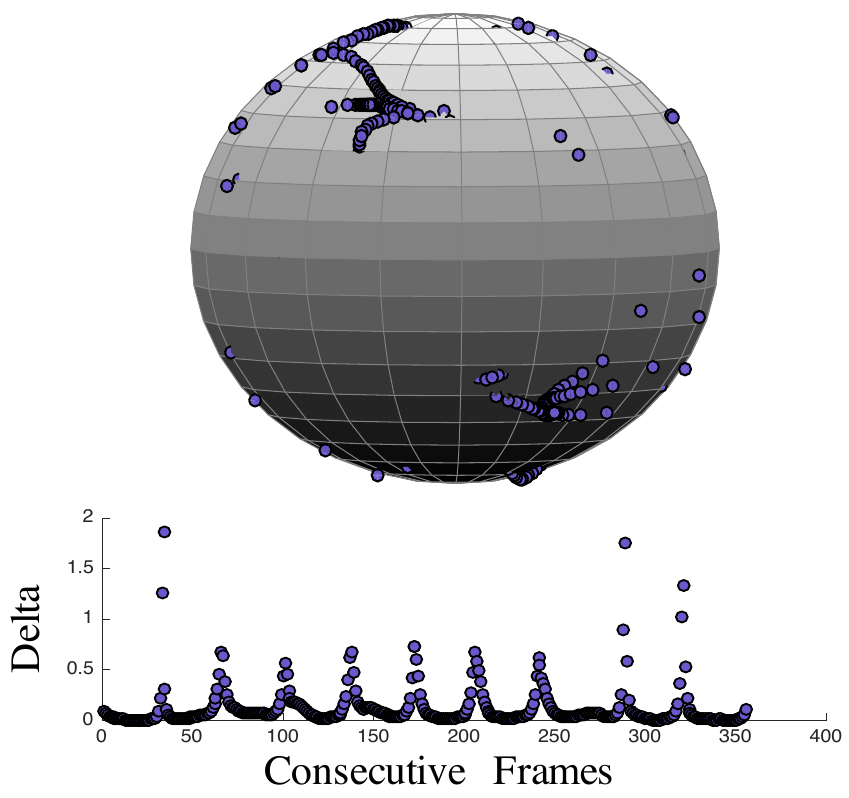}}}
\subfigure [$\m R \gets \m G_{12}$] {\label{fig:rot12} {\includegraphics[width=0.09\textwidth, height=0.09\textheight]{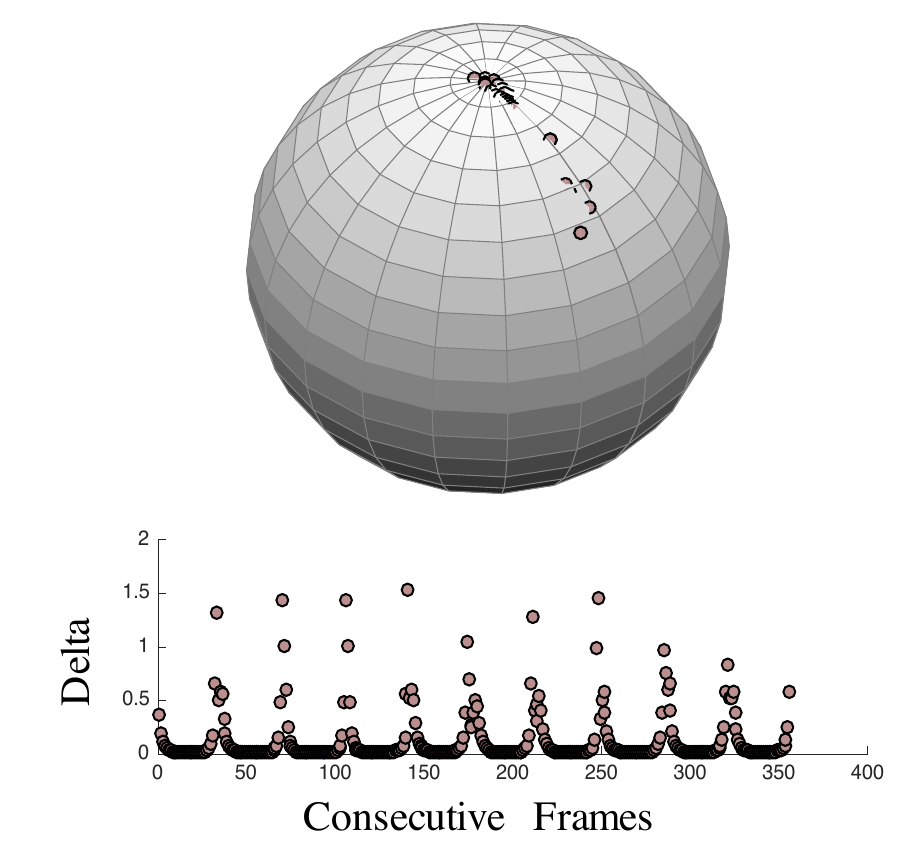}}}
\caption{\small{The rotation samples on $\mathbb{SO}(3)$ using $\{\m G_{\m k}\}_{\m k=1}^{12}$ for \emph{Pick-up sequence}. Below each $\mathbb{SO}(3)$ manifold is the graph showing the per frame change in the camera motion using Eq:(\ref{eq:rsmooth}) `$\delta_{\m f}$'. A simple observation establishes that all rotation matrix ($\m R $) are not the same. `$\delta_{\m f}$' graph analysis on this dataset show that the rotation estimate provided by $\m G_{7}, \m G_{8}$, $\m G_{9}$ has a smoother camera motion than other $\m G_{\m k}$'s, with $\m G_9$ being the smoothest. Any one out of these 3 $\m G_{\m k}$'s supply better performance than $\m G_{1}$. Note: Each $\m R_{\m i}\in \mathbb{R}^{2 \times 3} \mapsto \m R_{\m i}\in \mathbb{R}^{3 \times 3}$ via cross product. (Best viewed on screen)}}
\label{fig:RotationDemo}
\end{figure}

\begin{figure}
\centering
% \subfigure [\label{fig:a1}] {\includegraphics[width=0.13\textwidth, height=0.11\textheight]{Figures/EuclideanCorrective.pdf}}
\subfigure [\label{fig:a2}] {\includegraphics[width=0.220\textwidth, height=0.12\textheight]{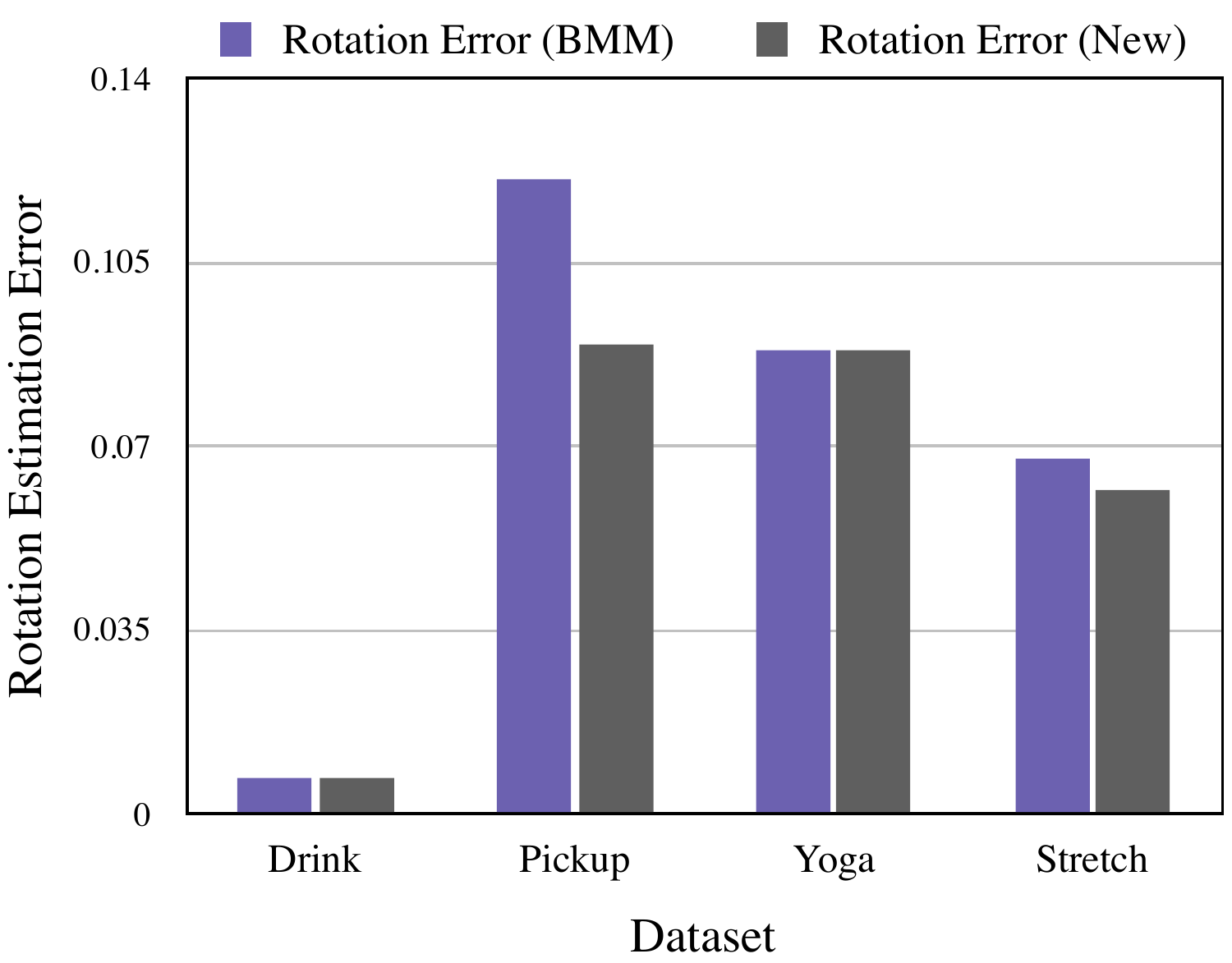}}
\subfigure [\label{fig:a3}] {\includegraphics[width=0.220\textwidth, height=0.12\textheight]{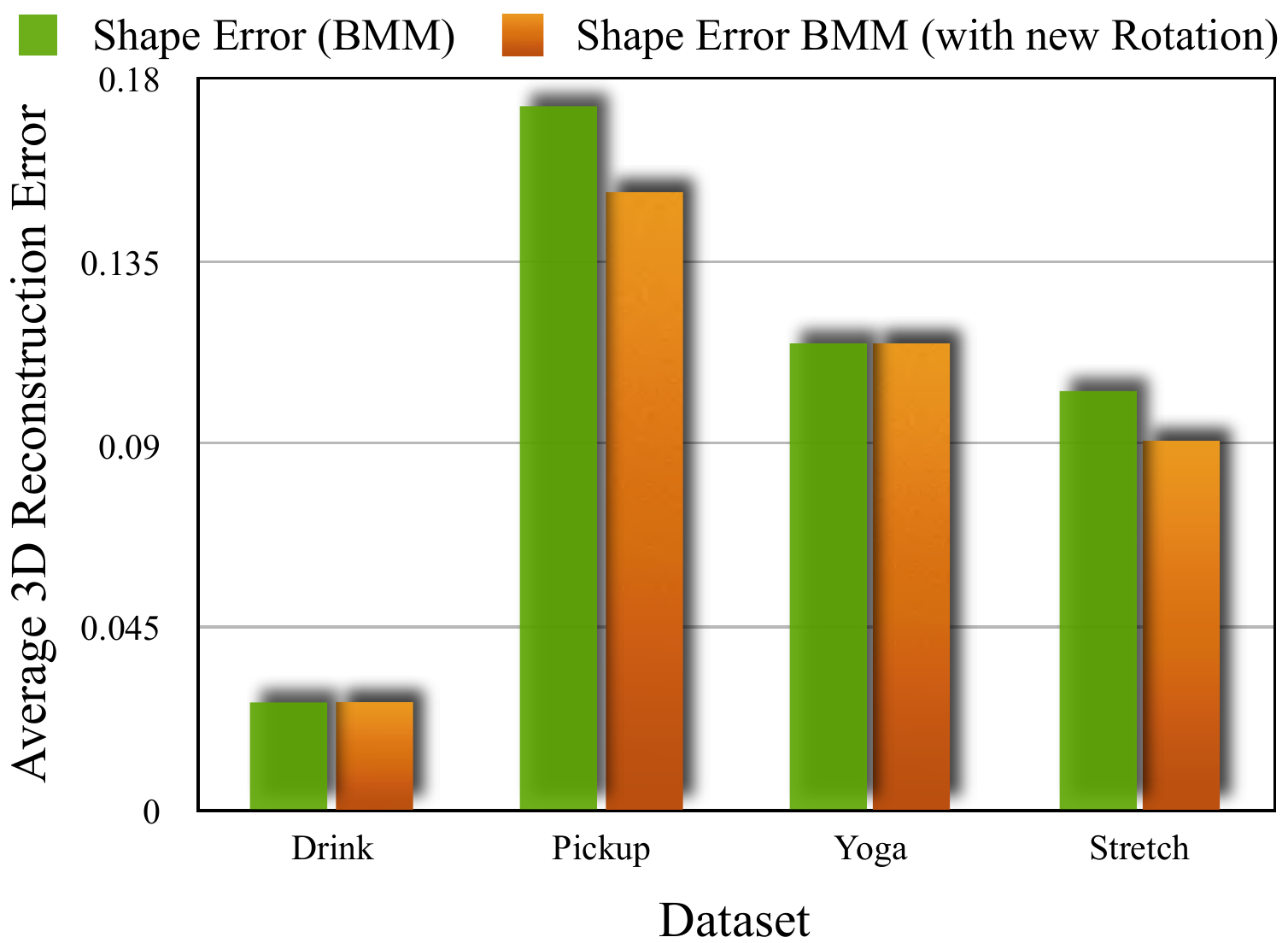}}
\caption{ \small{Counter examples on benchmark dataset \cite{akhter2011trajectory}. (a) Rotation error in comparison to BMM \cite{dai2014simple} on synthetic data. (b) 3D reconstruction error using global trace norm minimization of shape matrix as used in BMM with rotation matrix estimate using other column triplet in comparison to $\m G$(1:3). The column triplets of ($\m G$) for which the method perform better on Drink, Pickup, Yoga and Stretch are (1:3), (19:21), (1:3) and (19:21) respectively. Note that we used the same rank prior value `$\m K$' used in Dai \etal work \cite{dai2014simple}.}}
\label{fig:noiseandNsVariation}
\end{figure}

\noindent
{\bf{Plausible Rectification}}: Our experiments show that Dai \etal \cite{dai2014simple} solution to rotation estimation actually aborted the useful information present in the $\m G \in \mathbb{R}^{3 \m K \times 3 \m K}$. Each of the `$\m K$' column triplets in $\m G$ (\ie $\m G_{\m k}$) gives a possible rotation matrix which is different from each other (see Fig:(\ref{fig:RotationDemo})). Our empirical evaluations on several datasets show that the first column triplet is \textbf{not} always the best choice to estimate rotation. Hence, the details provided by Dai \etal work \cite{dai2014simple} is \textbf{incomplete} and there is a \emph{loss of generality} with such procedure to estimate rotation under the well-known assumption of smooth deformation \cite{rabaud2008re}. Fig:(\ref{fig:a2}) and Fig:(\ref{fig:a3}) provides few statistical results with comparison for both rotation and shape error estimate respectively. For clarity, we also provide the column triplet index that gives the better results for the corresponding data sequence and therefore, provides few counter-examples to such generalization.  

Theoretically and practically, this result is of significant importance 
as it helps in inferring that the solution provided by ``Intersection Theorem" has a lot of useful information left to be exploited completely and Dai \etal work ignored this. Also, it gives rise to some challenges that finding the best column triplet for $\m G_{\m k}$ is not an easy task. With these results, we \textbf{conjecture} few {problems} for further research in NRSfM that are: (a) Can we find a best possible column triplet for the corrective matrix with a given rank  prior `$(\m K)$', or (b) At least can we put an upper bound on the value $\m k \subset \m K$ such that there exists no such `$\m k$' for $\m G_{\m k}$ which will provide better rotation and structure estimate. The problem seems hard keeping in view that the prior rank $(\m K)$ in NRSfM factorization methods is an assumed approximation and it changes for different datasets to achieve better results.\\
\noindent
{\bf{A solution:}} In 2005, Brand. M \cite{brand2005direct} argued to use full correction matrix to estimate motion which in a way utilizes all the multiple estimates of column-triads of $\m G$. Recently, Lee \etal\cite{lee2016consensus} briefly mentions on the problems with motion estimates using \cite{dai2014simple}. In contrast, we use an analytical observation based on the smoothness and regularity\footnote{The term {{\textless\textless}}\textcolor{blue}{regularity}{{\textgreater\textgreater}} is used in a loose sense (Mathematically). } of the camera motion trajectory to filter $\m G_{\m k} \in \mathbb{R}^{3 \m K \times 3}$ to infer better `$\m R$'. Let $\psi(.)$ be a function that takes $\m G_{\m k}$ as input and gives `$\m R$' as output using Intersection theorem. We estimate different $\m R \in \mathbb{R}^{2\m F \times 3\m F}$ for all the column triplets $\{ \m G_{\m k}\}_{\m k = 1}^{\m K}$, then compute smoothness of the camera motion for each $\m G_{\m k}$ as:
\begin{equation}\label{eq:rsmooth}
\begin{aligned}
& \displaystyle \textrm{Suppose,} ~\m R = \psi(\m G_{\m k}), ~\textrm{via Intersection method, then, }\\
& \displaystyle \delta_{\m f} = \|\m R_{\m f} - \m R_{\m f+1}\|_{\m F}^{2} ~~\forall ~\m f=1, 2, ..., \m F-1. ~\cite{hartley2013rotation} ~\textrm{Sec.{\bf{4}}}. 
\end{aligned}
\end{equation}
By examining the smoothness of the camera motion for each $\m G_{\m k}$, we select the suitable rotation matrix for structure estimation (see Fig. \ref{fig:RotationDemo}).
Our strategy to select smooth camera motion over frames based on Eq:(\ref{eq:rsmooth}) consistently supplied us with better performance than the previously proposed approach. We acknowledge that this is not a profound way to infer the best rotation, however, it does provide a possibility to deduce better rotation using ``prior-free" approach which respects the well-known assumption of smooth deformation in NRSfM. Further, it helps endorse our claim on the generalization of rotation estimate by \cite{dai2014simple}. You may use the variable `$\delta_{\m f}$'  Eq:(\ref{eq:rsmooth}) as a smoothness term in the final optimization (Eq:(\ref{eq:10})) to further improve rotation, however, to show the competence within the ``prior-free'' idea \cite{dai2014simple}, we stick to the classical two staged approach.

\section{Structure Estimation}
Once the rotation is estimated based on the smoothness of the camera motion \cite{rabaud2008re}, the next step is to solve for 3D structure. The block matrix method (BMM) by Dai \etal \cite{dai2014simple} proposed the following optimization problem to estimate the non-rigid low-rank shape.
\begin{equation}\label{eq:8}
\begin{aligned}
& \displaystyle \underset{\m S^{\sharp}} {{\fontfamily{cmtt}\selectfont \text{minimize}}}~\|\m S^{\sharp}\|_{*} {\fontfamily{cmtt}\selectfont ~\text{subject to:}} ~\m W = \m R \m S, \m S^{\sharp} = g(\m S)
\end{aligned}
\end{equation}
where, $ \m S^{\sharp} \in \mathbb{R}^{\m F \times \m 3\m P}$ is a rearranged shape matrix with each row corresponds to the shape for that frame. The trace norm minimization on `$\m S^{\sharp}$' is enforced instead of `$\m S$' to provide a stronger rank bound on the shape matrix \cite{dai2014simple}. The second term in Eq:(\ref{eq:8}) enforces the re-projection error constraint. The function $g(.)$ maps $\m S \in \mathbb{R}^{3 \m F \times \m P}$ to $\m S^{\sharp} \in \mathbb{R}^{\m F \times 3 \m P}$.

\noindent
{\bf{Dai \etal solution to shape}}: Following the work of Ma \etal \cite{ma2011fixed} on rank minimization problems, Dai \etal \cite{dai2014simple} proposed a solution to the optimization in Eq:(\ref{eq:8}). The method enforces low-rank constraint on `$\m S^{\sharp}$' matrix and provide the solution by solving Eq:(\ref{eq:8}) via ADMM\cite{boyd2011distributed} using matrix shrinkage operator $\mathcal{S}_{\lambda}(\m X)$ = $\m U \text{diag}(s_{\lambda}(\sigma)) \m V^{\m T}$, where $s_{\lambda}(\sigma)$ = $\bar{\sigma}$ with $\bar{\sigma_{i}}$ = $\big\{ \sigma_i -\lambda ~\text{if} ~\sigma_i -\lambda > 0$ and 0 otherwise \big\}.

\noindent
{\bf{Plausible Rectification}}: Despite the trace norm minimization provides a satisfactory solution to non-rigid structure estimation, it has some serious issues. The proposed solution to nuclear norm minimization problem (Eq:\eqref{eq:8}) gives equal priority to each singular values, as a result, the shrinkage operator penalizes each singular value with the same quantity ($\lambda$). To estimate 3D structure of a non-rigidly deforming object using matrix factorization approach, we use a prior assumption that the shape lies in a low-rank subspace. Therefore, it's not a better choice to penalize the major component of the shape data and its very minor component equally. Consequently, nuclear norm minimization of the shape matrix struggles to appropriately conserve the useful component of the non-rigidly deforming shape.

%fails to successfully exploit the prior knowledge about the rank of the non-rigid deforming shapes. 

Truncated nuclear norm regularization can be a choice to handle such issues, however, it depends on the binary decision, hence not versatile in nature \cite{zhang2012matrix}. To really cater the behavior of the deformations based on its low-rank nature, we propose to use weighted nuclear norm minimization approach to solve for non-rigid structure \cite{srebro2003weighted,gu2014weighted}. In contrast to the previous notation to the nuclear norm of the shape matrix \ie $\| \m S^{\sharp}\|_*$, we introduce a different notation for its weighted nuclear norm
\begin{equation}
\| \m S^{\sharp}\|_{\Theta, *} = \sum_{\m j = 1}^{\m K} \Theta_{\m j}\sigma_{\m j}(\m S^{\sharp}) 
\end{equation}

where $\sigma_{\m j}(.)$ denotes the $\m j^\textrm{th}$ singular value of $\m S^{\sharp}$. We assume that the weights $\Theta_{\m j}$'s are non-negative scalar \ie $\Theta_{\m j} \geq 0$ . Using this representation, we redefine the optimization proposed in the Eq:(\ref{eq:8}) as follows:

\begin{equation}\label{eq:10}
\begin{aligned}
& \displaystyle \underset{\m S^{\sharp}, \m S} {\fontfamily{cmtt}\selectfont {\text{minimize}}} ~ \mu \|\m S^{\sharp}\|_{\Theta, *} + \frac{1}{2}\| \m W - \m R \m S\|_{\m F}^{2}\\
& \displaystyle {\fontfamily{cmtt}\selectfont \text{subject to:}}~ \m S^{\sharp} = g(\m S)
\end{aligned}
\end{equation}

The motivation for such formulation is quite clear, however, the proposed optimization (Eq:\ref{eq:10}) is generally \textbf{non-convex}, and is more difficult to solve than the nuclear norm minimization. Fortunately, recent results \cite{zha2017analyzing, lu2015nonconvex, gu2014weighted} in compressed sensing have shown that we can achieve an effective optimal solution to Eq:(\ref{eq:10}) in the case when $0 \leq \Theta_1 \leq \Theta_{2} \leq .... \leq \Theta_{\m K}$ \S \ref{ss:opt}.

\subsection{Optimization}\label{ss:opt}
\noindent
This section provides the mathematical derivation to the optimization proposed in Eq:(\ref{eq:10}). Our solution use the following theorems and proofs as stated and used in \cite{zha2017analyzing, gu2014weighted,cai2010singular}. 

\begin{theorem}
For all $\m Y \in \mathbb{R}^{\m m \times \m n}$, denoted by $\m Y = \m U \Sigma \m V^{\m T}$, the SVD of it. The solution to $\textrm{minimize}_{\m X}\| \m Y - \m X\|_{\m F}^{2} + \| \m X\|_{\Theta, *}$, with non-negative weight vector $\Theta$, its solution $\hat{\m X}$ can be written as $\hat{\m X} = \m U \hat{\m B} \m V^{\m T}$, where $\hat{\m B}$ is the solution to the following optimization problem
\begin{equation}
\hat{\m B} = \textrm{argmin}_{\m B} \|\Sigma - \m B\|_{\m F}^{2} + \| \m B \|_{\Theta, *}
\end{equation}
\end{theorem}

%\begin{theorem}
%If the weights satisfy $\Theta_1 \geq \Theta_{2} \geq ... \Theta_{\m K} \geq 0$, the weighted nuclear norm minimization problem $\textrm{minimize}_{\m X}\| \m Y - \m X\|_{\m F}^{2} + \| \m X\|_{\Theta, *}$ has a globally optimal solution
%\begin{equation}
%\hat{\m X} = \m U \mathcal{S}_{\Theta}(\Sigma) \m V^{\m T}
%\end{equation}
%where $\m Y = \m U \Sigma \m V^{\m T}$ is the SVD of $\m Y$, and $\mathcal{S}_{\Theta}(\Sigma)$ is the generalized soft-thresholding operator with weight vector $\Theta$
%\begin{equation}
%\mathcal{S}_{\Theta}(\Sigma) = \text{max}(\Sigma_{\m i \m i}-\Theta_{\m i})
%\end{equation}
%\end{theorem}

\begin{theorem}
If the singular values $\sigma_1 \geq .... \geq \sigma_{\m K}$ and the weights satisfy $0 \leq \Theta_1 \leq \Theta_{2} \leq .... \leq \Theta_{\m K}$ then 
the weighted nuclear norm minimization problem $\textrm{minimize}_{\m X}\| \m Y - \m X\|_{\m F}^{2} + \| \m X\|_{\Theta, *}$ has a globally optimal solution
\begin{equation}
\hat{\m X} = \m U \mathcal{S}_{\Theta}(\Sigma) \m V^{\m T}
\end{equation}
where $\m Y = \m U \Sigma \m V^{\m T}$ is the SVD of $\m Y$, and $\mathcal{S}_{\Theta}(\Sigma)$ is the generalized soft-thresholding operator with weight vector $\Theta$
\begin{equation}
\mathcal{S}_{\Theta}(\Sigma) = \text{max}(\Sigma_{\m i \m i}-\Theta_{\m i}, 0)
\end{equation}
\end{theorem}

\noindent
The readers are encouraged to refer to \cite{zha2017analyzing, gu2014weighted} work for detailed derivations to the lemma's leading to the proof of the theorems. In conclusion, if the weights satisfies non-descending order, not necessarily with the same value, the weighted nuclear norm minimization problem is still convex and optimal solution can be obtained using a soft-thresholding operator with different weights \cite{zha2017analyzing, gu2014weighted}.

\subsection{Solution}
We propose our solution to the optimization problem defined in Eq:(\ref{eq:10}) using alternating direction method of multipliers \cite{boyd2011distributed} (ADMM), a simple, fast but powerful algorithm used to solve many {non-convex} problems in computer vision and mathematical optimization. The ADMM algorithm decompose the original problem into several sub-problems, where each of them is solved separately by introducing Lagrange multipliers and penalty parameters to estimate convergence. Using the method of multipliers, the Augmented Lagrangian form for Eq:(\ref{eq:10}) is written as follows:
\begin{equation}\label{eq:14}
\begin{aligned}
& \displaystyle \mathcal{L}_{\rho}(\m S^{\sharp}, \m S) = \mu \|\m S^{\sharp}\|_{\Theta, *} + \frac{1}{2}\| \m W - \m R \m S\|_{\m F}^{2} + \frac{\rho}{2}\|\m S^{\sharp} - g(\m S)\|_{\m F}^2 + \\
& \displaystyle ~~~~~~~~~~~~~~~~~~~<\m Y, \m S^{\sharp}-g(\m S)>
\end{aligned}
\end{equation}
here $\m Y \in \mathbb{R}^{\m F \times 3 \m P}$ is a Lagrange multiplier and $\rho > 0$ is the penalty parameter. The solution to each variable is obtained by solving the following subproblems over iterations (indexed with the variable $\m i$):
\begin{equation}\label{eq:15}
\begin{aligned}
& \displaystyle (\m S^{\sharp})^{\m i+1} = \underset{\m S^{\sharp}}{{\fontfamily{cmtt}\selectfont \text{argmin}}} ~\mathcal{L}_{\rho}\big((\m S^{\sharp})^{\m i}, \m S\big)\\
% & \displaystyle = \underset{\m S^{\sharp}}{\text{argmin}} ~\mu \|\m S^{\sharp}\|_{\Theta, *} + \frac{\rho}{2}\|\m S^{\sharp} - g(\m S)\|_{\m F}^2 + <\m Y, \m S^{\sharp}-g(\m S)>
\end{aligned}
\end{equation}

\begin{equation}\label{eq:16}
\begin{aligned}
& \displaystyle(\m S)^{\m i+1} = \underset{\m S}{{\fontfamily{cmtt}\selectfont \text{argmin}}} ~\mathcal{L}_{\rho}\big(\m S^{\sharp}, (\m S)^{\m i}\big)\\
% & \displaystyle = \underset{\m S}{\text{argmin}} \frac{\| \m W - \m R \m S\|_{\m F}^{2}}{2} + \frac{\rho\|\m S^{\sharp} - g(\m S)\|_{\m F}^2}{2} +<\m Y, \m S^{\sharp}-g(\m S)>
% & \displaystyle = \big(\frac{\rho\m I + \m R^{\m T}{\m R}}{\rho}\big) \big \backslash \Big( \big(g^{-1}(\m S^{\sharp}) + \frac{g^{-1}(\m Y)}{\rho} \big) + \frac{\m R^{\m T}\m W}{\rho} \Big)
\end{aligned}
\end{equation}
The Lagrange multiplier and the penalty parameter are updated as follows:
\begin{equation}
\begin{aligned}
& \displaystyle \m Y = \m Y + \rho(\m S^{\sharp} - g(\m S))\\
& \displaystyle \rho = {\fontfamily{cmtt}\selectfont \text{minimum}}(\rho_\textrm{max}, \lambda\rho)
\end{aligned}
\end{equation}
$\rho_\textrm{max}$ refers to the maximum value of `$\rho$' and $\lambda$ is an empirical constant ($\lambda>1$). The mathematical derivations to each sub-problems are provided in the supplementary material for reference. The closed form solution to the Eq:(\ref{eq:16}) is obtained by taking the derivative of Eq:(\ref{eq:14}) w.r.t variable `$\m S$' and equating it to zero \ie,
\begin{equation}
\begin{aligned}
& \displaystyle \m S = \Big(\frac{\rho\m I + \m R^{\m T}{\m R}}{\rho}\Big)^{-1} \Big( \big(g^{-1}(\m S^{\sharp}) + \frac{g^{-1}(\m Y)}{\rho} \big) + \frac{\m R^{\m T}\m W}{\rho} \Big)
\end{aligned}
\end{equation}
%Note `$\backslash$' is a Matlab slang \ie if $\m A \m x = \m B$ implies $\m x = \m A\backslash \m B$. 
Similarly, rewriting the Eq:(\ref{eq:14}) treating $\m S^{\sharp}$ as variable.

\begin{equation}\label{eq:19}
\begin{aligned}
\displaystyle = \underset{\m S^{\sharp}}{{\fontfamily{cmtt}\selectfont \text{argmin}}} ~\mu \|\m S^{\sharp}\|_{\Theta, *} + \frac{\rho}{2}\|\m S^{\sharp} - g(\m S)\|_{\m F}^2 +<\m Y, \m S^{\sharp}-g(\m S)>
\end{aligned}
\end{equation}
In contrast to the previous form, the solution to Eq:(\ref{eq:19}) is not straight forward. To obtain a closed form solution to this problem, lets define a soft-thresholding function $\mathcal{S}_{\tau}(\m \sigma) = \textrm{sign}(\sigma).\textrm{max}(|\sigma|-\tau, 0)$. Also,  let $[\m U, \Sigma, \m V]$ be the singular value decomposition of ($g(\m S)-\frac{\m Y}{\rho}$), then the optimal solution to Eq:(\ref{eq:19}) is given by:
\begin{equation}
\m S^{\sharp} = \m U \mathcal{S}_{\frac{\Theta\mu}{\rho}}(\Sigma)\m V
\end{equation}
Here, $\Theta$ is the weight assigned to the different singular values in the non-descending order based on its significance to the deformation data. For detail discussion on the initialization of weights kindly refer section \S \ref{ss:winit}.

% Mathematically, its hard to put an upper bound on the number of columns triplets that is required to estimate best correction matrix in the factorization approach \cite{bregler2000recovering}. Nevertheless, Dai {\it{et al.}} formulation \cite{dai2014simple} neglected the best column triplet present within the estimated correction matrix, hence the word ``uncertainty".
\section{Experiment and Discussion}\label{sec:EandD}
To endorse our claim, we performed extensive experiments on both real and synthetic benchmark datasets \cite{akhter2009nonrigid,jensen2018benchmark,torresani2008nonrigid}. We compared the performance of our algorithm against different state-of-the-art methods on these datasets \cite{gotardo2011non,lee2013procrustean,kumar2017spatio}. Additionally, we unveil the substantial percentage boost in the reconstruction accuracy as high as 18\% in comparison to the previous results reported for ``simple prior-free" approach. For real-world applications to NRSfM, noisy data and missing feature tracks over frames are crucial, therefore, we also performed experiments to tackle such issues. To make the comparisons on noisy and missing data sequence, the experimental settings we used are  same and consistent with Lee \etal work \cite{lee2013procrustean}. Experimental results on dense datasets and more rigorous cases of missing trajectories are provided in the supplementary material. Before we provide details on our performance analysis, lets discuss the variable initialization.
\subsection{Initialization}
Our algorithm has few parameters and variables to initialize. For all our experiments on different datasets, we initialize $\mu=1$, $\lambda = 1.1$, $\rho_\textrm{max} = 1e^{10}$, $\rho = 1e^{-4}$, $\m Y = \textrm{zeros}(\m F, 3 \m P)$ and the `$\m K$' values are kept same as Dai \etal method \cite{dai2014simple}. Practically, we considered the convergence of our optimization, if the gap  ${\fontfamily{cmtt}\selectfont \text{max}}\|(\m S^{\sharp} - g(\m S))\|_{\infty}<1e^{-8}$ or $\rho >\rho_\textrm{max}$ over iteration.  
% \begin{figure*}
% \centering
% \includegraphics[width=1\textwidth] {Figures/CMUResult.pdf}~~~
% \caption{\small{Reconstruction results of our method on the NRSfM synthetic benchmark dataset \cite{akhter2009nonrigid, akhter2011trajectory}. Ground-truth and reconstructed points are shown in filled and non-filled circles respectively.}}
% \label{fig:CMUResult}
% \end{figure*}

\begin{figure*}
\centering
\subfigure[Shark]{\label{fig:nr1} {\includegraphics[width=0.20\textwidth, height=0.120\textheight]{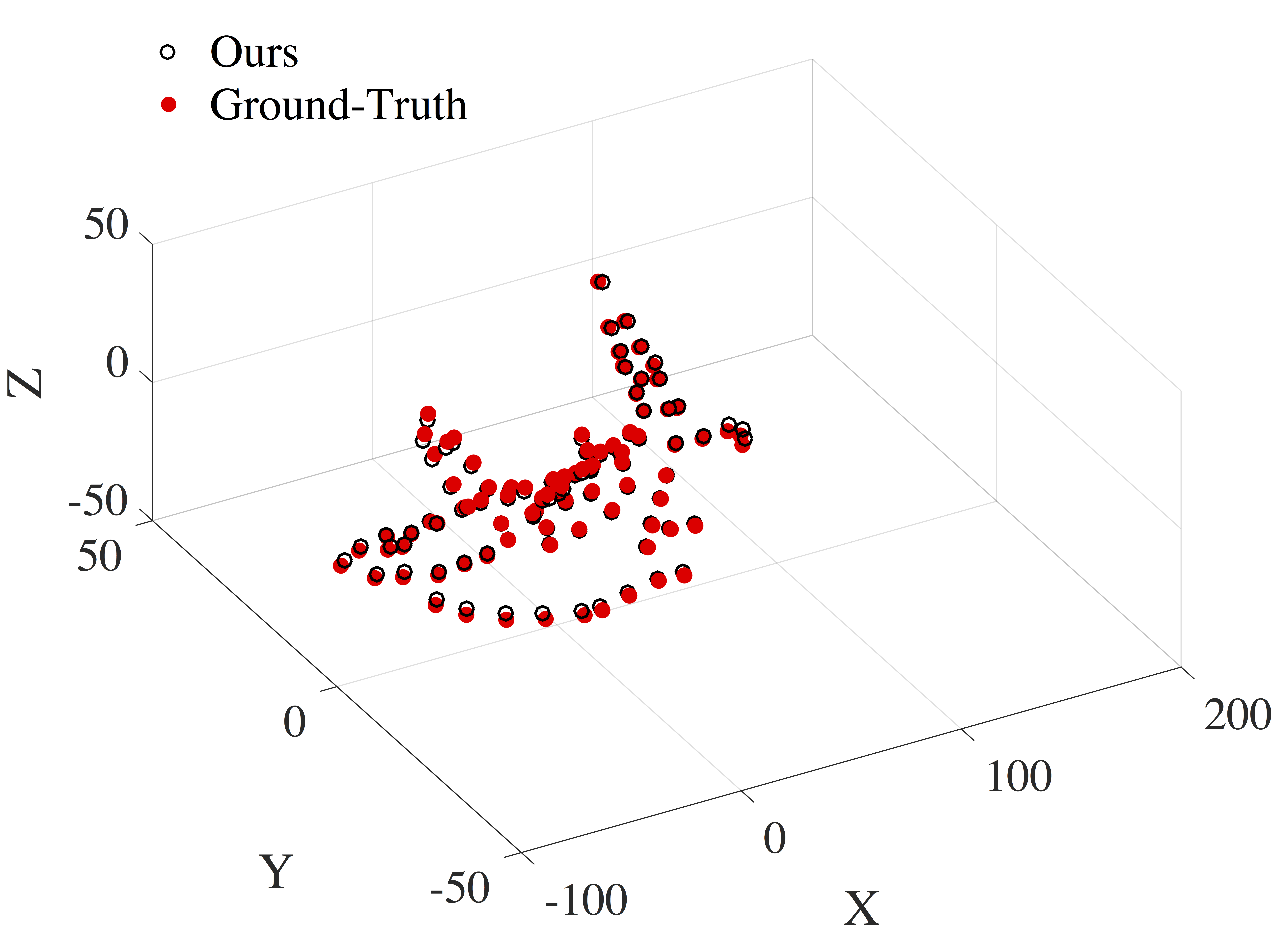}}}
\subfigure[Pick-up]{\label{fig:nr2}{\includegraphics[width=0.20\textwidth, height=0.120\textheight]{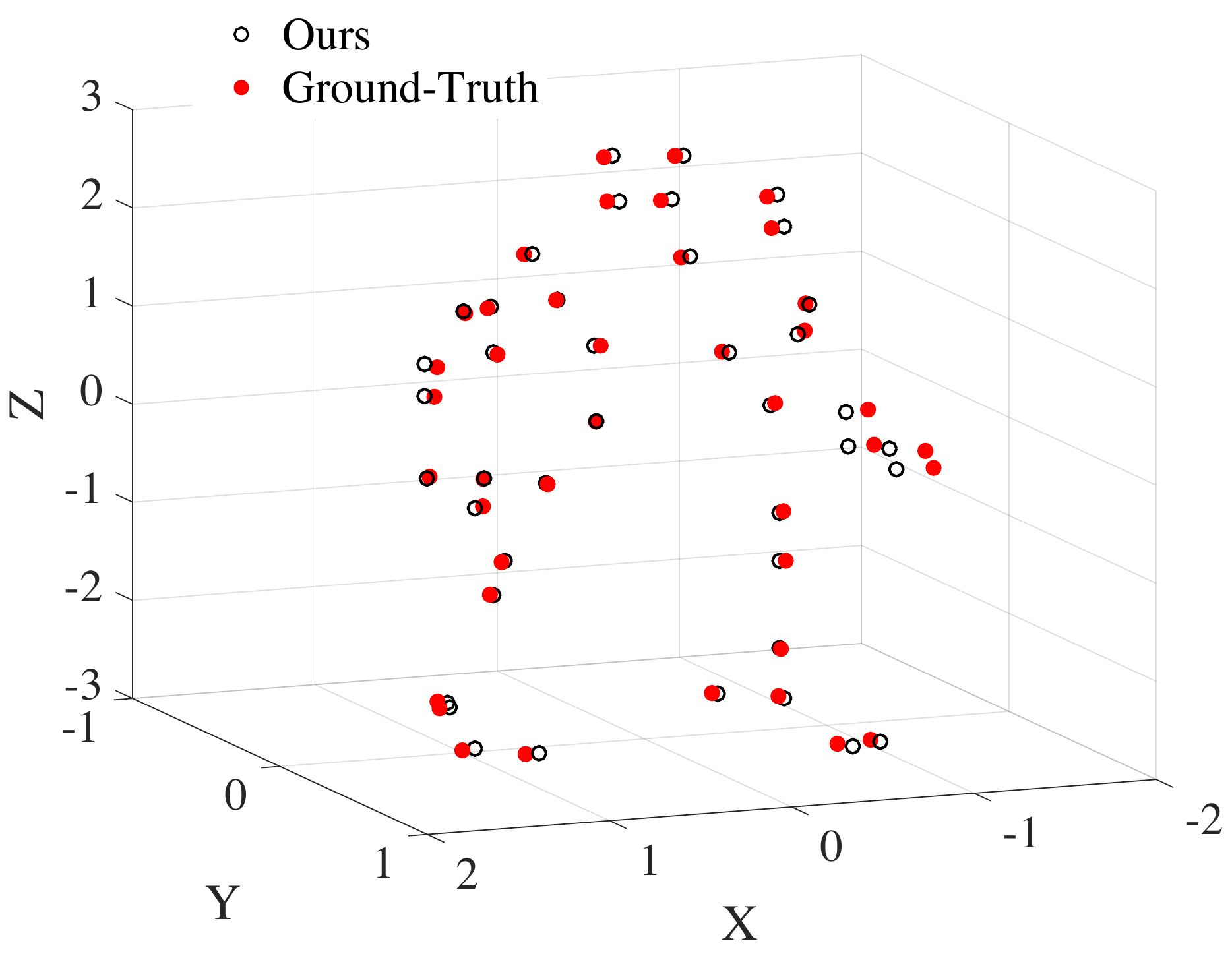}}}
\subfigure[Drink]{\label{fig:nr3} {\includegraphics[width=0.20\textwidth, height=0.120\textheight]{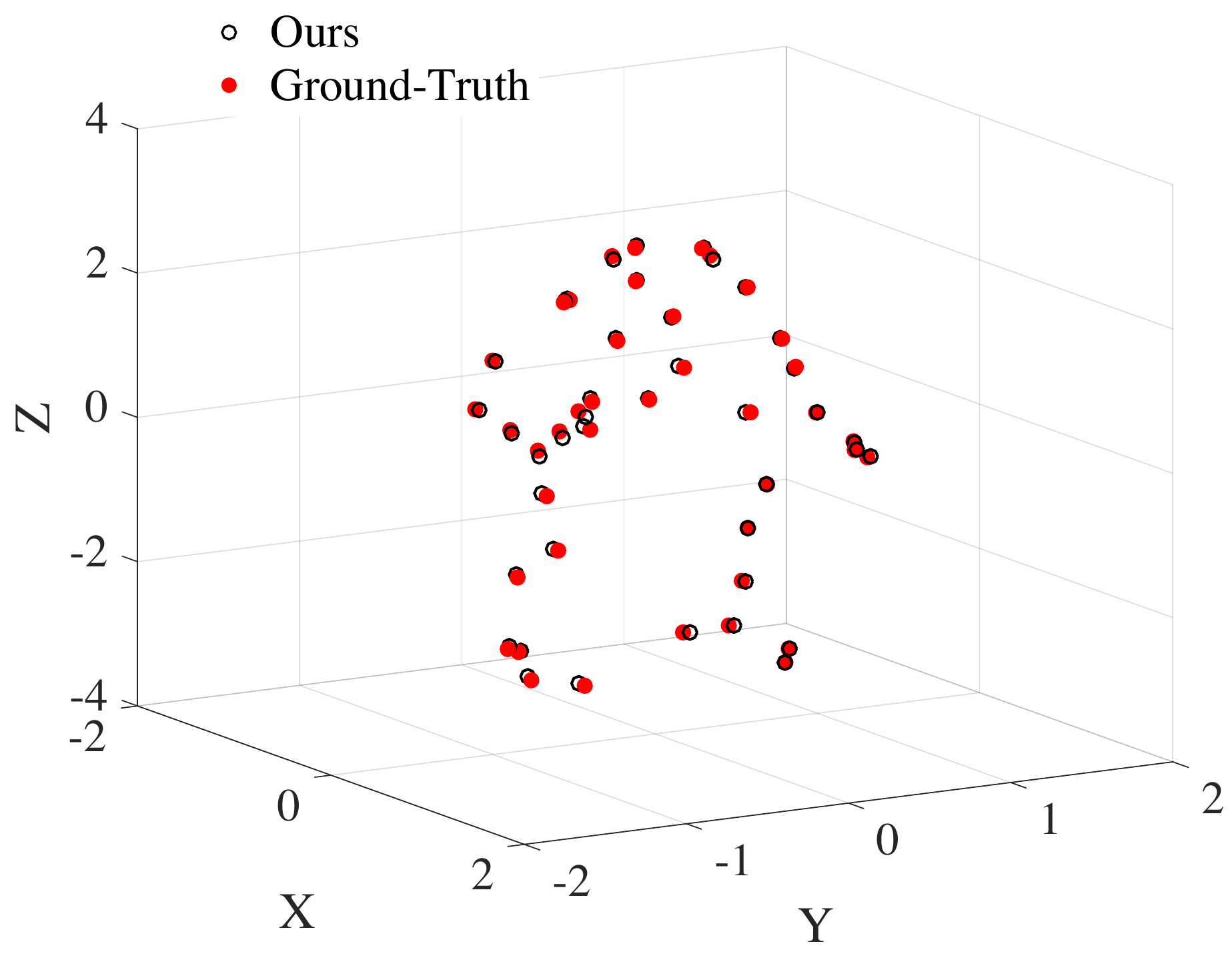}}}
\subfigure[Face] {\label{fig:nr4}{\includegraphics[width=0.20\textwidth, height=0.120\textheight]{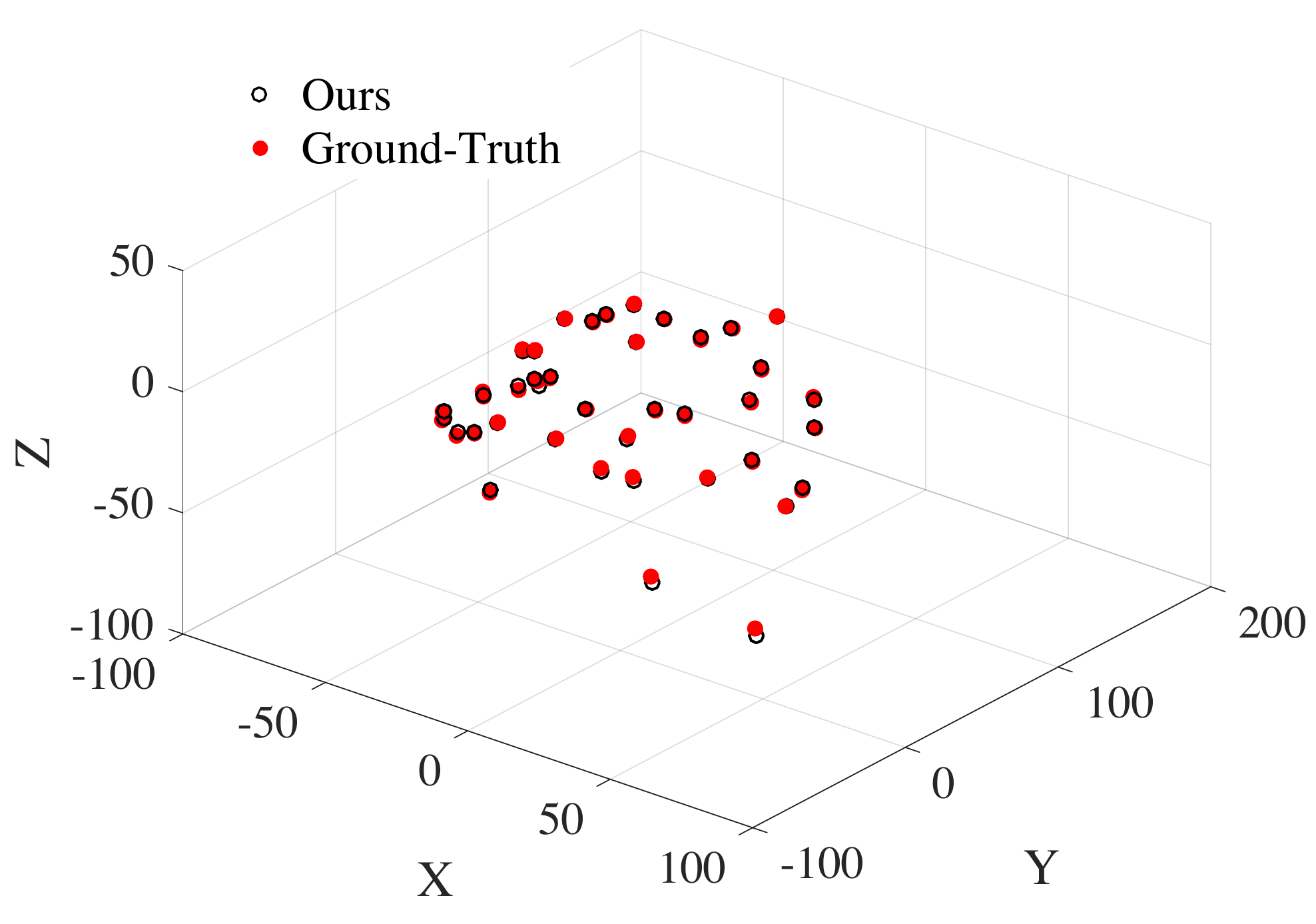}}}

\subfigure[Dance] {\label{fig:nr5} {\includegraphics[width=0.20\textwidth, height=0.120\textheight]{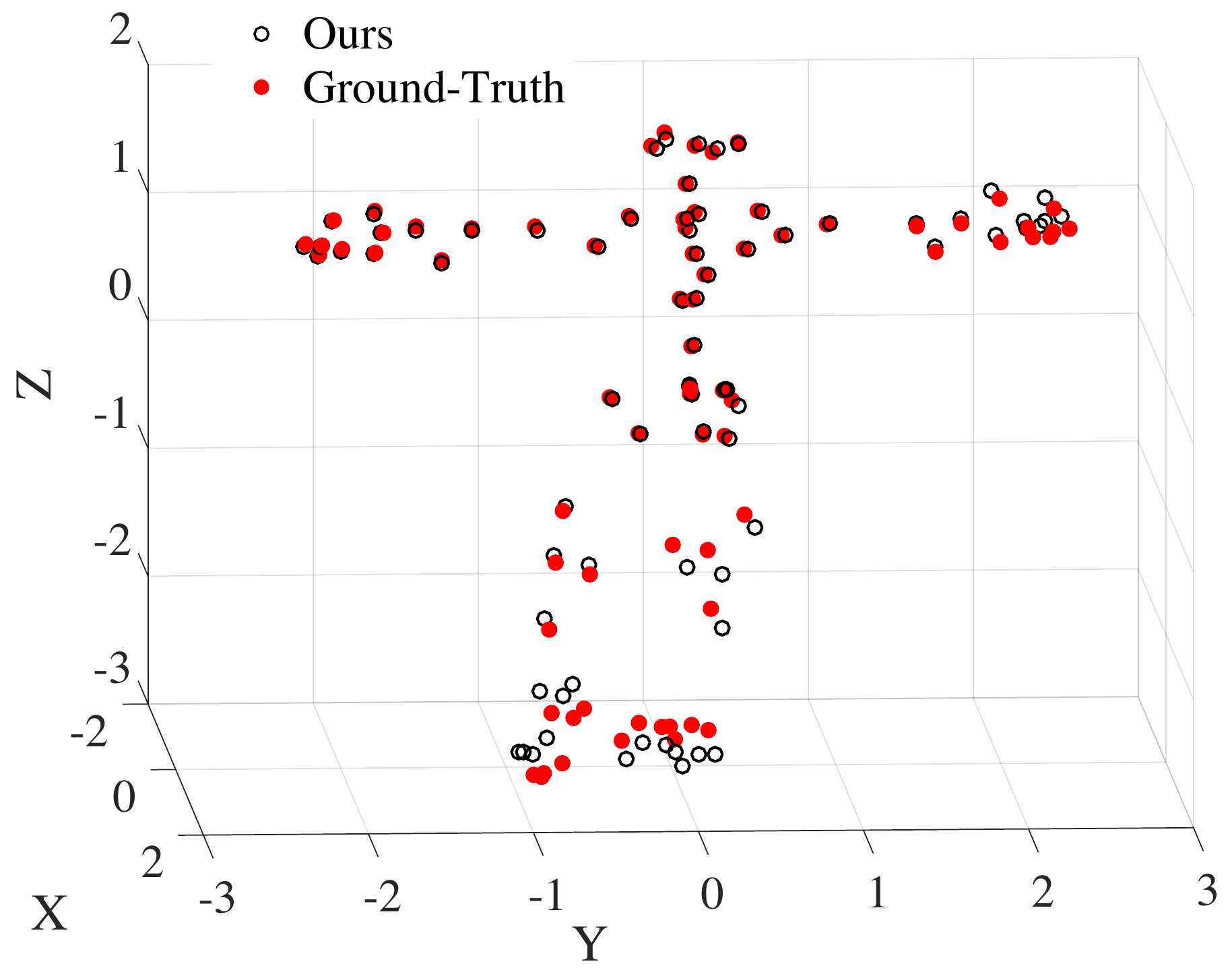}}}
\subfigure[Yoga] {\label{fig:nr6} {\includegraphics[width=0.20\textwidth, height=0.120\textheight]{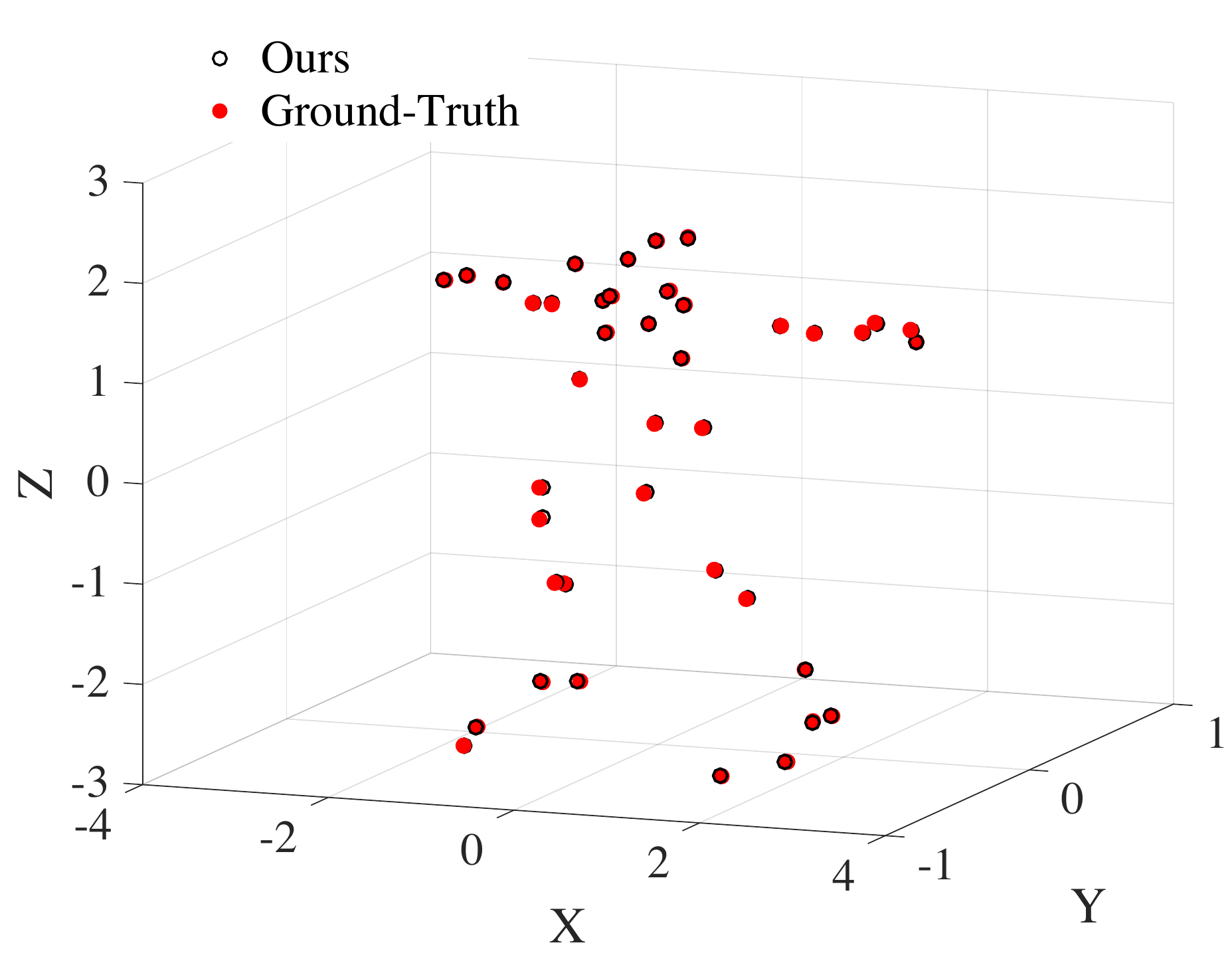}}}
\subfigure[Walking] {\label{fig:nr7}{\includegraphics[width=0.20\textwidth, height=0.120\textheight]{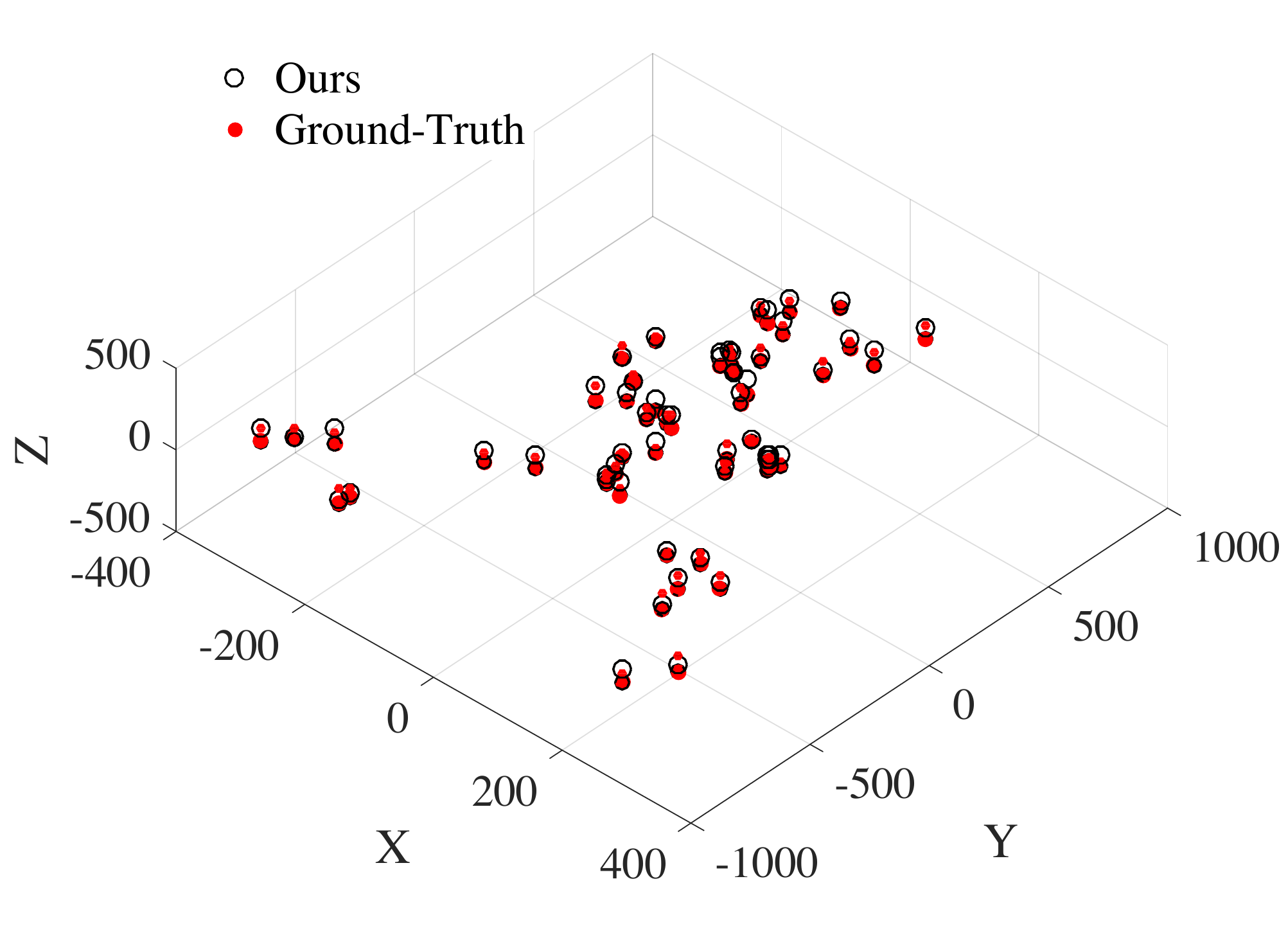}}}
\subfigure [Stretch] {\label{fig:nr8} {\includegraphics[width=0.20\textwidth, height=0.120\textheight]{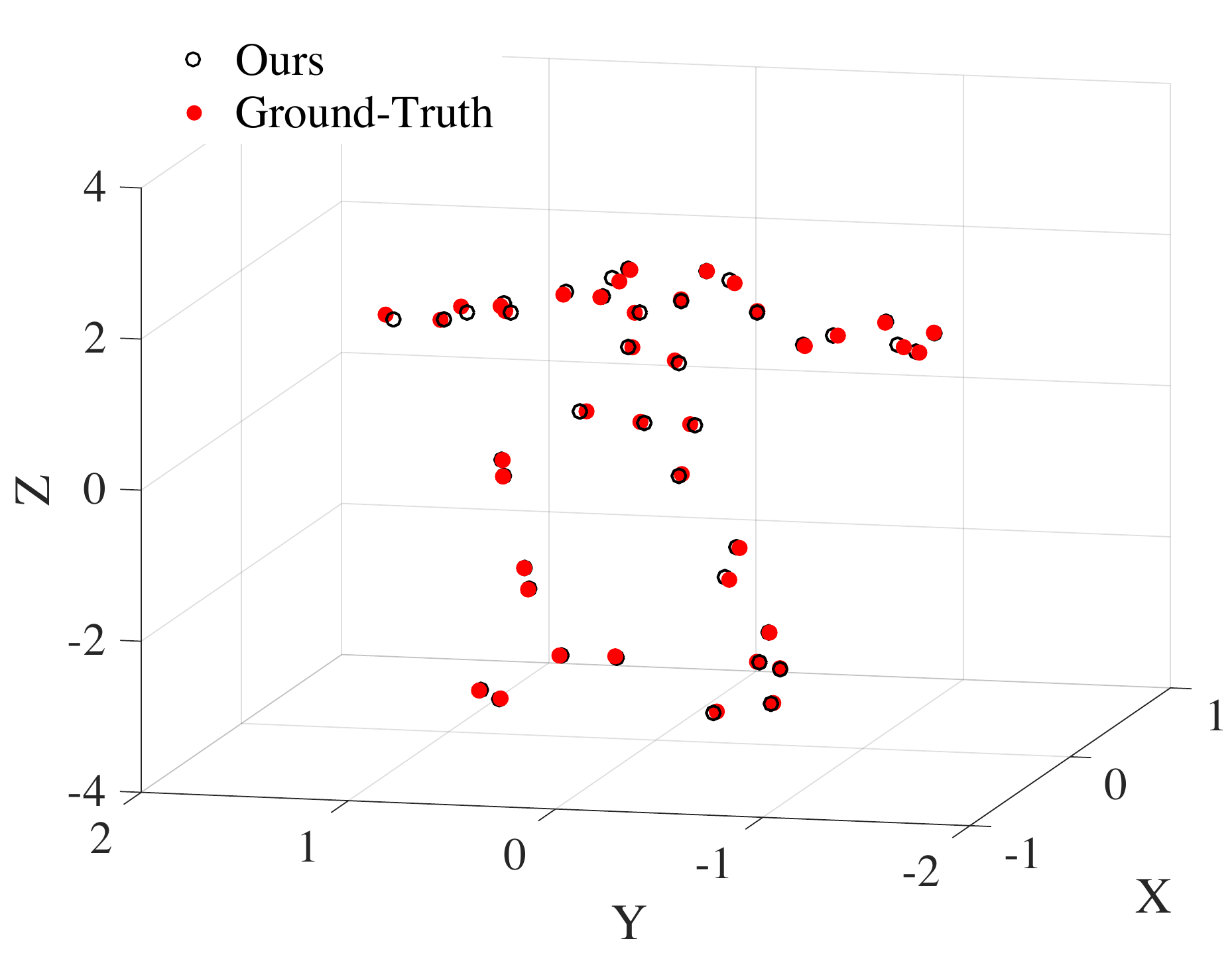}}}
\caption{\small{Reconstruction results of our method on the NRSfM synthetic benchmark dataset \cite{akhter2009nonrigid, akhter2011trajectory}. Ground-truth and reconstructed points are shown in filled(red) and non-filled circles respectively. Note: We used same `$\m K$' value as documented in \cite{dai2014simple} work for all the experiments.}}
\label{fig:CMUResult}
\end{figure*}

\noindent
{\bf{1. Structure initialization:}}\label{ss:sinit}
Using the result of Liu \etal \cite{liu2013robust} on the uniqueness of minimizer for the rank minimization problem, we initialize the the 3D shape `$\m S$' as `$\m S$' = ${\fontfamily{cmtt}\selectfont \text{pinv}}(\m R)\m W$ and $\m S^{\sharp} = g(\m S)$. The pseudo-inverse solution to shape matrix provides a good enough initialization to our algorithm. Reader may refer to Dai \etal \cite{dai2014simple} and Valmadre \etal \cite{valmadre2015closed} work for detailed discussion on pseudo inverse solution to `$\m S$' in NRSfM.\\
\noindent
{\bf{2. Weight ($\Theta$) initialization:}}\label{ss:winit}
It is well-known in NRSfM under factorization approach that the shape matrix lies in a low-rank space. Generally, the largest singular value of the shape matrix contains the most information about the non-rigid shape, therefore, while optimizing for the shape matrix, it's illogical to treat each singular value equally.  The singular values with major component must be penalized less and vice-versa. Using this inverse relation between singular values and its significance to the shape deformation modeling, we assign the weight ($\Theta$) to be inversely propositional to the singular values of the shape matrix.
\begin{equation}\label{eq:21}
\Theta_{\m j} = \frac{\xi}{\sigma_{\m j}(\m S^{\sharp}) + \gamma} = \frac{\xi}{\sigma_{\m j}(g(\m S)) + \gamma} 
\end{equation}
where, $\xi$ is a positive number and $\gamma = 1e^{-6}$, a very small positive number to avoid division by zero as some singular values are likely to be zero (low rank). We initialized the weights by substituting the pseudo-inverse initialization of `$\m S^{\sharp}$'  \ie using the relation $\m S^{\sharp}$ = $g(\m S)$ in the Eq:(\ref{eq:21}).

\subsection{Performance Analysis}
\begin{figure}
\centering
\subfigure[Articulated]{\label{fig:nrc1} {\includegraphics[width=0.15\textwidth, height=0.09\textheight]{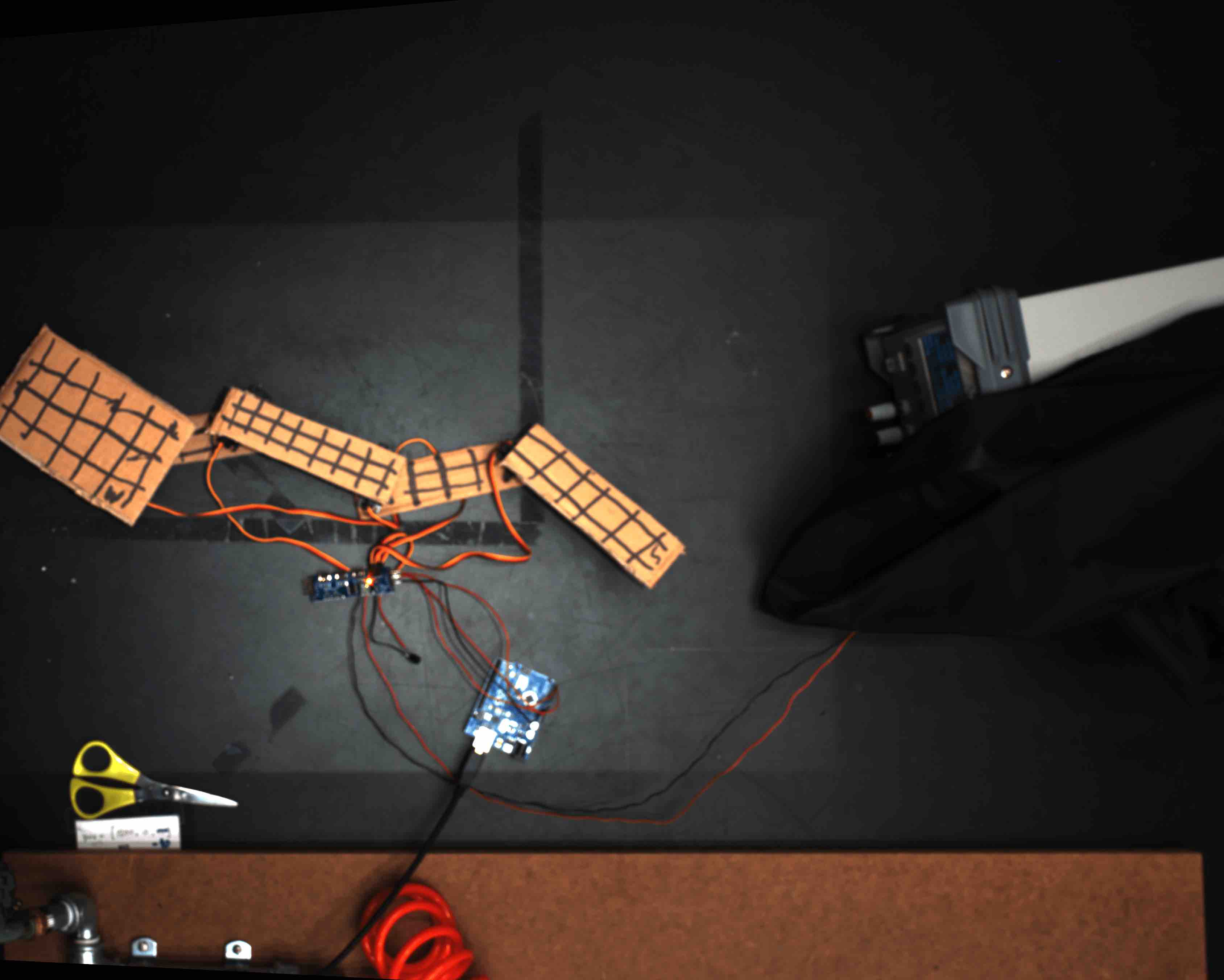}}}
\subfigure[Ballon]{\label{fig:nrc2}{\includegraphics[width=0.15\textwidth, height=0.09\textheight]{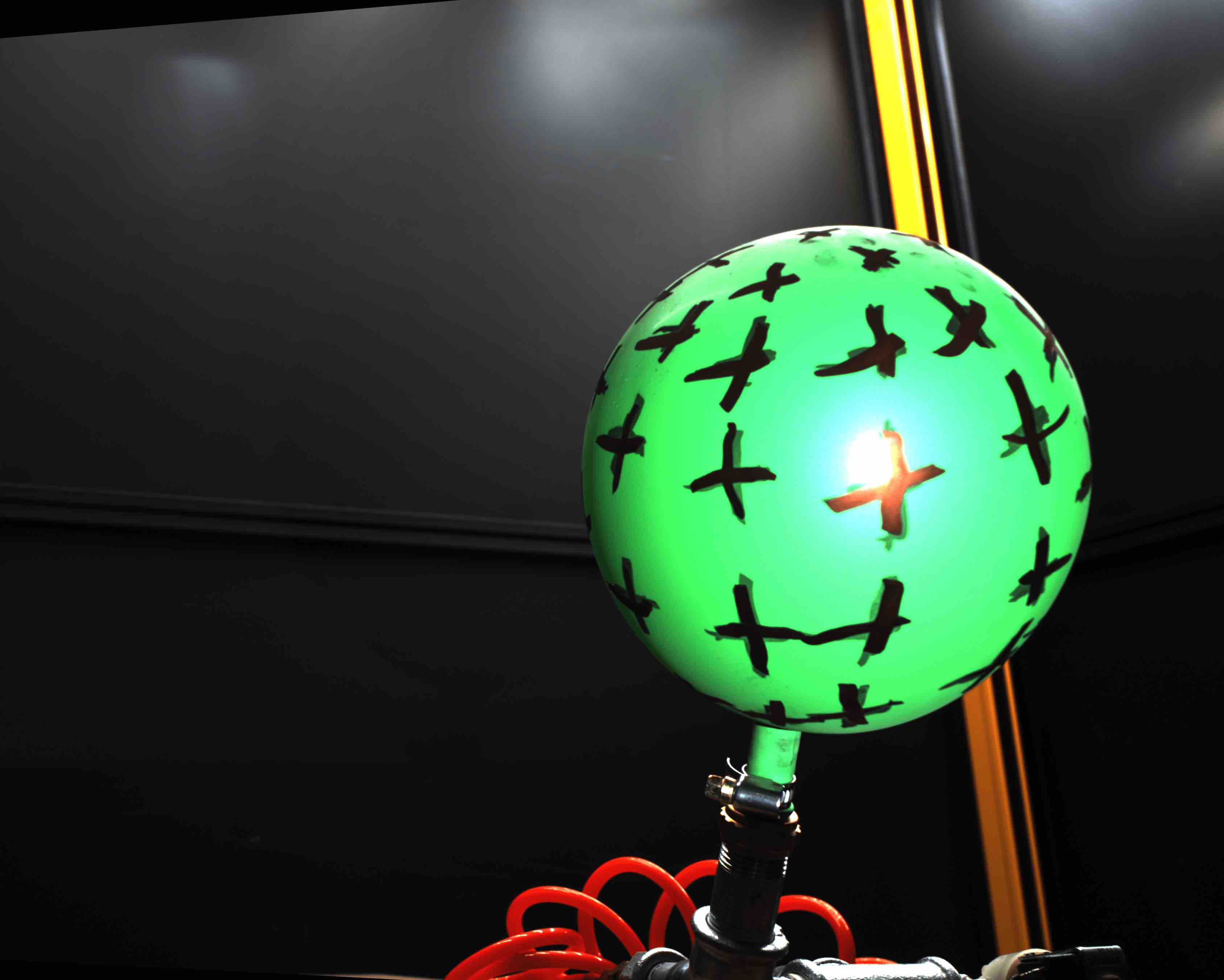}}}
\subfigure[Stretch]{\label{fig:nrc3} {\includegraphics[width=0.15\textwidth, height=0.09\textheight]{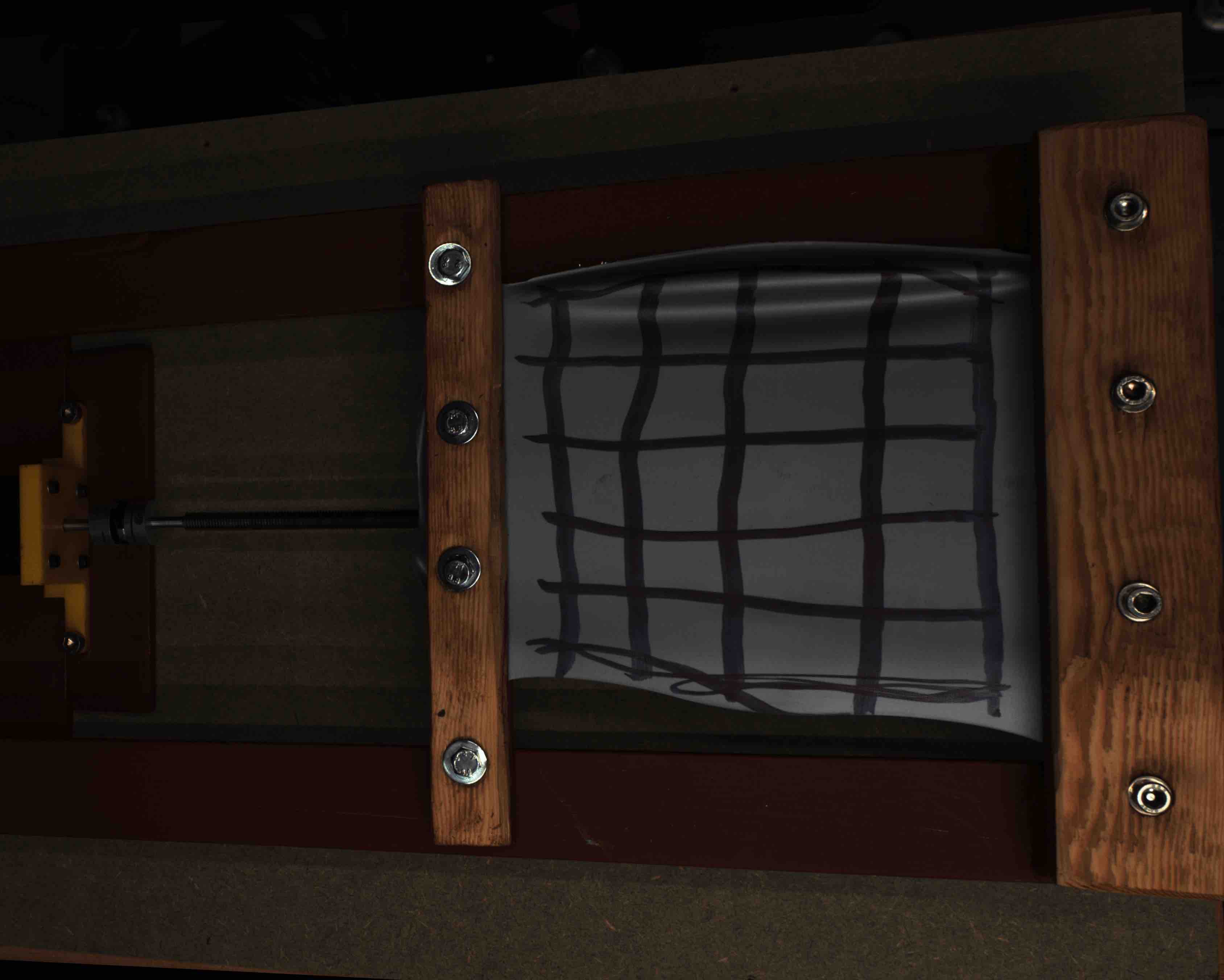}}}

\subfigure[3D Articulated] {\label{fig:nrc4} {\includegraphics[width=0.15\textwidth, height=0.09\textheight]{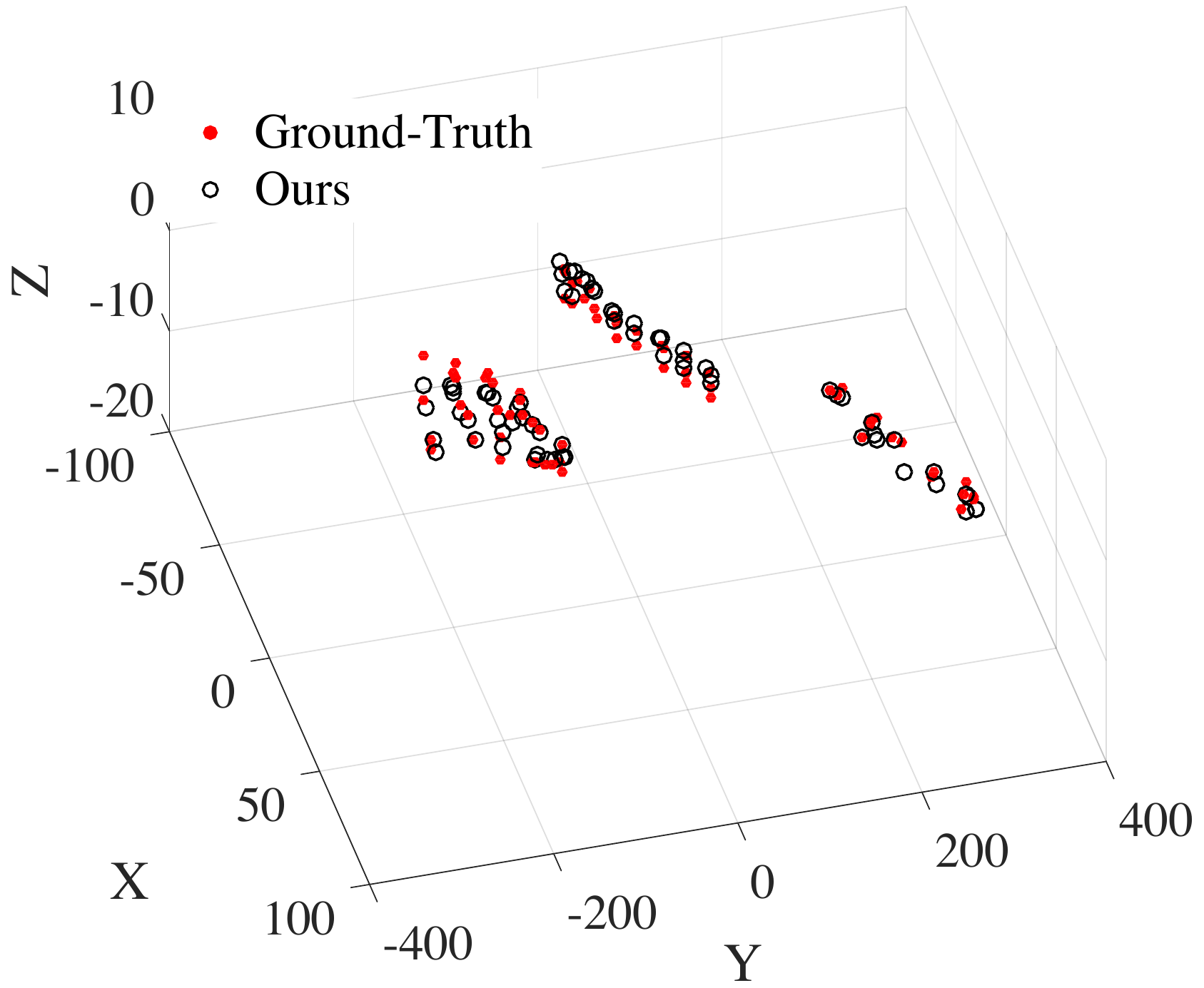}}}
\subfigure[3D Ballon] {\label{fig:nrc5} {\includegraphics[width=0.15\textwidth, height=0.09\textheight]{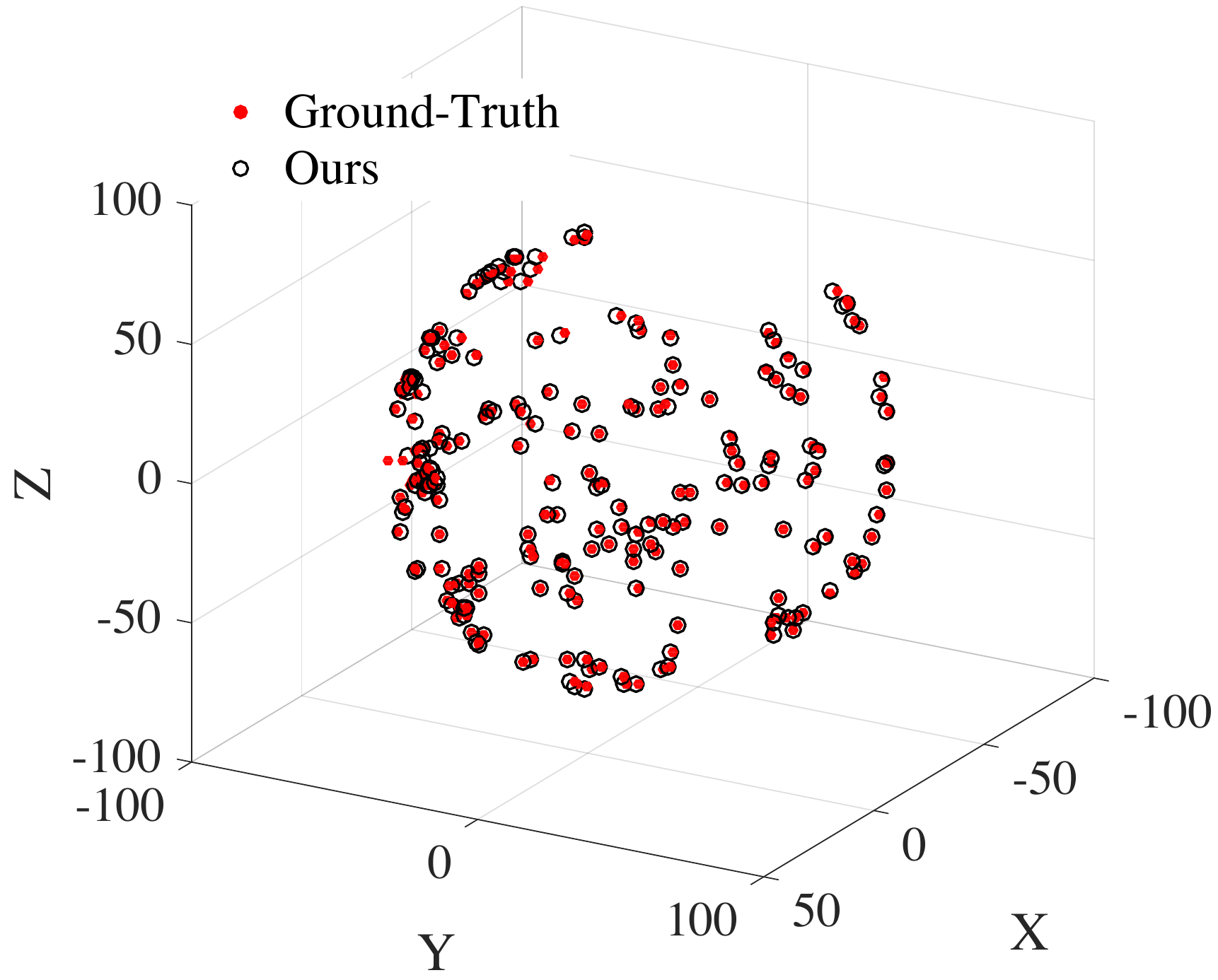}}}
\subfigure[3D Stretch] {\label{fig:nrc6}{\includegraphics[width=0.15\textwidth, height=0.09\textheight]{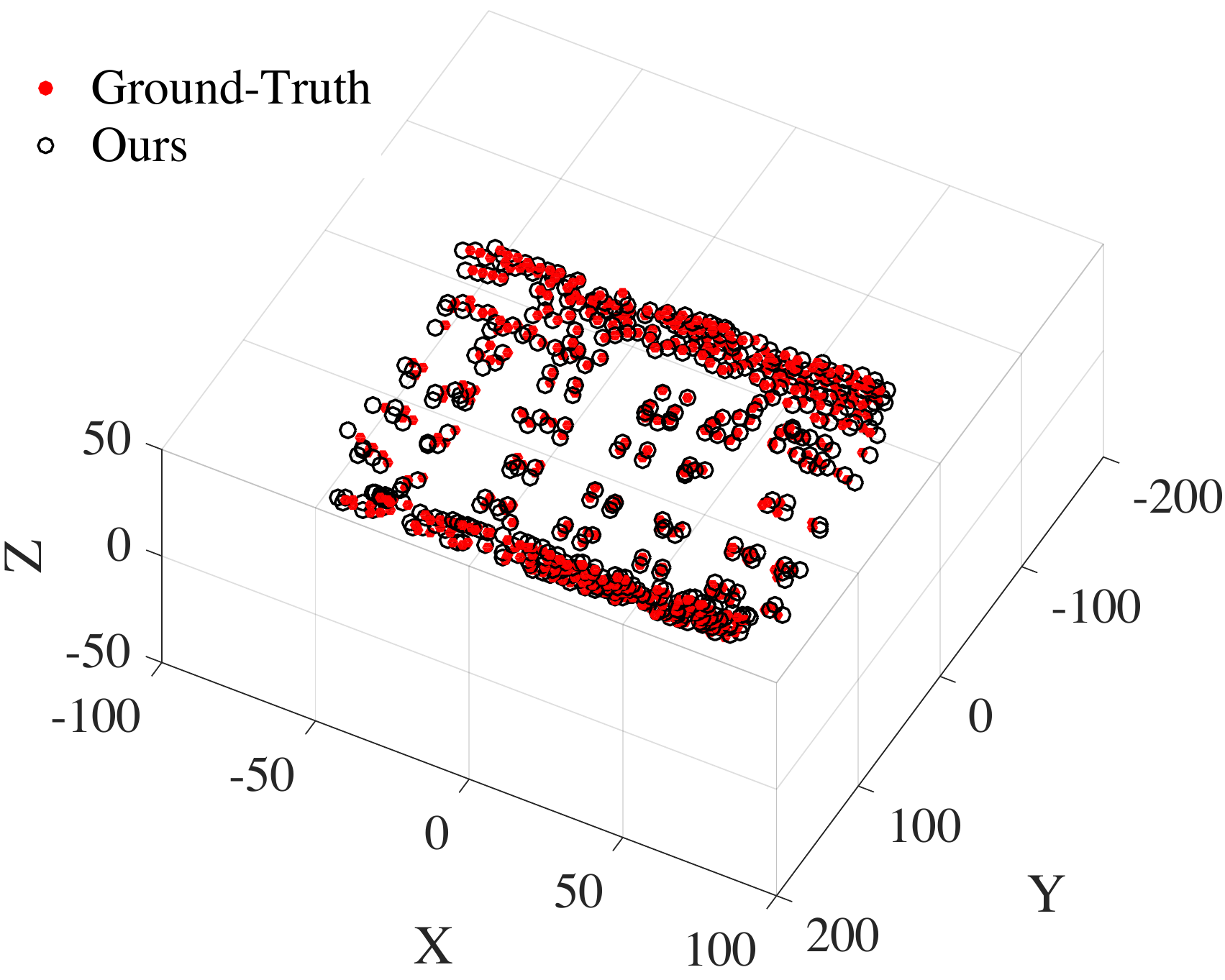}}}
\caption{\small{Reconstruction results of our method on the NRSfM challenge dataset \cite{jensen2018benchmark}. The results shown here are for the circular camera path. Ground-truth 3D and reconstructed 3D points are shown with filled and non-filled circles respectively.}}
\label{fig:challengeResult1}
\end{figure}

After a detailed discussion on the variable initialization and optimization, we present our experimental evaluation. We performed extensive experiments on both new and old benchmark datasets \cite{akhter2009nonrigid,torresani2008nonrigid,jensen2018benchmark}. We report the quantitative result on the previous benchmark dataset using mean normalized 3D reconstruction error formulation \ie
\begin{equation}
e_{s} = \frac{1}{\m F} \sum_{\m i = 1}^{\m F}\frac{\|\m S_\textrm{ est}^{\m i} - \m S_{\m G\m T}^{\m i}\|_{\m F}}{\|\m S_{\m G\m T}^{\m i}\|_{\m F}}
\end{equation}
where, $\m S_\textrm{ est}$, $\m S_{\m G \m T}$ are the estimated 3D shape and ground-truth 3D shape respectively. To keep our statistics consistent with the newly proposed NRSfM dataset, we used their error evaluation code to compute the robust root mean square error (RMSE) metric as proposed in Taylor \etal work \cite{taylor2010non}. For more details on NRSfM CVPR 2017 challenge dataset evaluation metric, kindly refer to Jensen \etal work \cite{jensen2018benchmark}.
\begin{table}
\centering
\scriptsize
\begin{tabular}{|>{\columncolor[gray]{0.88}}l|l|l|l|l|l|}
\hline
\rowcolor[gray]{0.75}
{\scriptsize{Method}} & {\scriptsize{PTA}}\cite{akhter2009nonrigid}  & {\scriptsize{CSF2}}\cite{gotardo2011non} & {\scriptsize{PND}}\cite{lee2013procrustean} & {\scriptsize{BMM}} & {\scriptsize{Ours}} \\ \hline
Drink  &  0.0287   &  0.0227   &  {\bf{0.0037}}   & 0.0266  &   0.0119 (\textcolor{red}{1.47\%})\\ \hline
Pickup &  0.1939   &  0.1791   &  0.0372   & 0.1731  &   {\bf{0.0198}} (\textcolor{red}{15.3\%})\\ \hline
Yoga   &  0.1243   &  0.1179   &  0.0140   & 0.1150  &   {\bf{0.0129}} (\textcolor{red}{10.2\%})\\ \hline
Stretch & 0.1035   &  0.1136   &  0.0156   & 0.1034  &   {\bf{0.0144}} (\textcolor{red}{8.90\%})\\ \hline
Dance   & 0.2426   &  0.1877   &  0.1454   & 0.1864  &   {\bf{0.1060}} (\textcolor{red}{8.04\%})\\ \hline
Walking & 0.3761   &  0.1938   & {\bf{0.0465}}   & 0.1298  &   0.0882 (\textcolor{red}{4.16\%})\\ \hline
Face   &  0.0489   &  0.0319   & {\bf{0.0165}}   & 0.0303  &   0.0179 (\textcolor{red}{1.24\%})\\ \hline
Shark  &  0.2933   &  0.1117   & {\bf{0.0135}}   & 0.2311  &   0.0551 (\textcolor{red}{17.6\%})\\ \hline
\end{tabular}
\caption{ \small{Performance comparison in the shape recovery using our new approach with some of the state-of-the-art methods in single body NRSfM. The statistics clearly demonstrate our claim that we can achieve a significant improvement in the reconstruction accuracy without using complex mathematical formulation. The percentage value in the last column (red) show the improvements over the result documented  by Dai \etal original work (BMM) \cite{dai2014simple}.}} \label{tab:statisticalresults1}
\end{table}
\\
\noindent
{\bf{1. Benchmark datasets:}}
Most of the methods proposed in non-rigid structure from motion often use it to evaluate the performance of the algorithm. Loosely speaking, this dataset is composed of eight standard sequences namely Drink, Pickup, Yoga, Stretch, Dance, Walking, Face and Shark. The number of frames ($\m F$) to number of points ($\m P$) \ie $(\m F, \m P)$ set for these datasets are (1102, 41), (357, 41), (307, 41), (370, 41), (264, 75), (316, 40) and (240, 91) respectively. Table (\ref{tab:statisticalresults1}) show the statistical comparison of our approach in comparison to the other competing approaches for single body NRSfM. Our evaluation results clearly show a significant improvement in the reconstruction accuracy in comparison to the previously reported results for ``prior-free'' approach. Figure (\ref{fig:CMUResult}) show the qualitative reconstruction results w.r.t ground-truth on all of these sequences.\\
\noindent
{\bf{2. NRSfM challenge datasets:}}
Jensen \etal recently released this dataset as a part of NRSfM competition held at CVPR 2017 \cite{jensen2018benchmark}. This is a high quality challenging dataset divided into five categories based on the deformation type, namely, Articulated, Balloon, Paper, Stretch and Tearing. Each of these categories is again shot using six different camera paths namely circle, flyby, line, semi-circle, tricky and zig-zag. This dataset is significantly larger and diverse to really test the performance of a NRSfM algorithm's. However, the dataset provides only a single frame ground-truth 3D for each of the five categories to test the algorithm. To estimate the reliability of our approach, we compared our performance against the best performing algorithm on this dataset.
Table (\ref{tab:statisticalresults2}) show the quantitative results of our method. The performance clearly demonstrates the significant improvement in the accuracy using ``prior-free'' idea under our modification. It also help infer that without using complex mathematical notions, we can reach performance accuracy close to the state-of-the-art. Figure (\ref{fig:challengeResult1}) show some qualitative results using our method. 
%Due to poor performance of recent method \cite{lee2016consensus} on this dataset \cite{jensen2018benchmark}, we avoided its discussion and comparison to our method.

\begin{table}
\centering
\scriptsize
\begin{tabular}{|>{\columncolor[gray]{0.88}}c|c|c|c|c|c|}
\hline
\rowcolor[gray]{0.75}
{\scriptsize{Method} $\downarrow$ / \scriptsize{Data}} & {\scriptsize{Articulated}}  & {\scriptsize{Ballon}} & {\scriptsize{Paper}} & {\scriptsize{Stretch}} & {\scriptsize{Tearing}} \\ \hline
Multibody \cite{kumar2017spatio} &  10.15   &  10.64   &  15.78   & 9.96  &   14.17\\ \hline
BMM \cite{dai2014simple}  &  24.54   &  12.91   &  22.37  & 18.71  &   18.87 \\ \hline
Ours  & 12.02   &  11.79   &  16.21   & 12.05  &   16.08 \\ \hline
\end{tabular}
\caption{ \small{Performance comparison of our method in comparison to the best performing algorithm (Multi-body) \cite{kumar2017spatio} on NRSfM challenge dataset \cite{jensen2018benchmark}. The above statistics shows the average root-mean-square error in millimeters for the single test image on the orthogonal sequence available with the dataset. Our method shows a clear improvement over the originally proposed BMM approach and it's accuracy got very close to the multi-body.}} \label{tab:statisticalresults2}
\end{table}

\begin{figure}
\centering
\subfigure [\label{fig:n1}] 
{\includegraphics[width=0.20\textwidth, height=0.11\textheight]{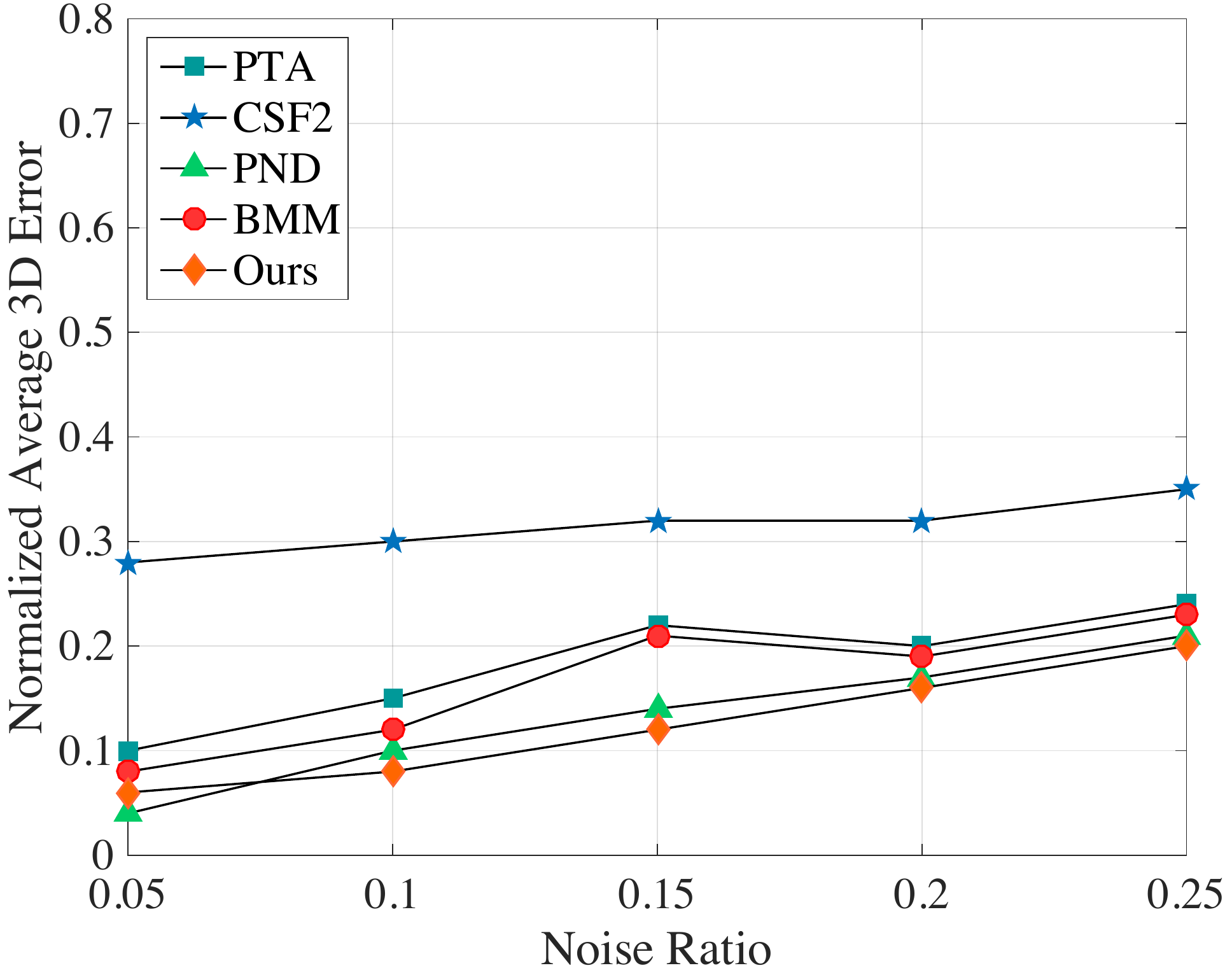}}
\subfigure [\label{fig:m1}] {\includegraphics[width=0.20\textwidth, height=0.11\textheight]{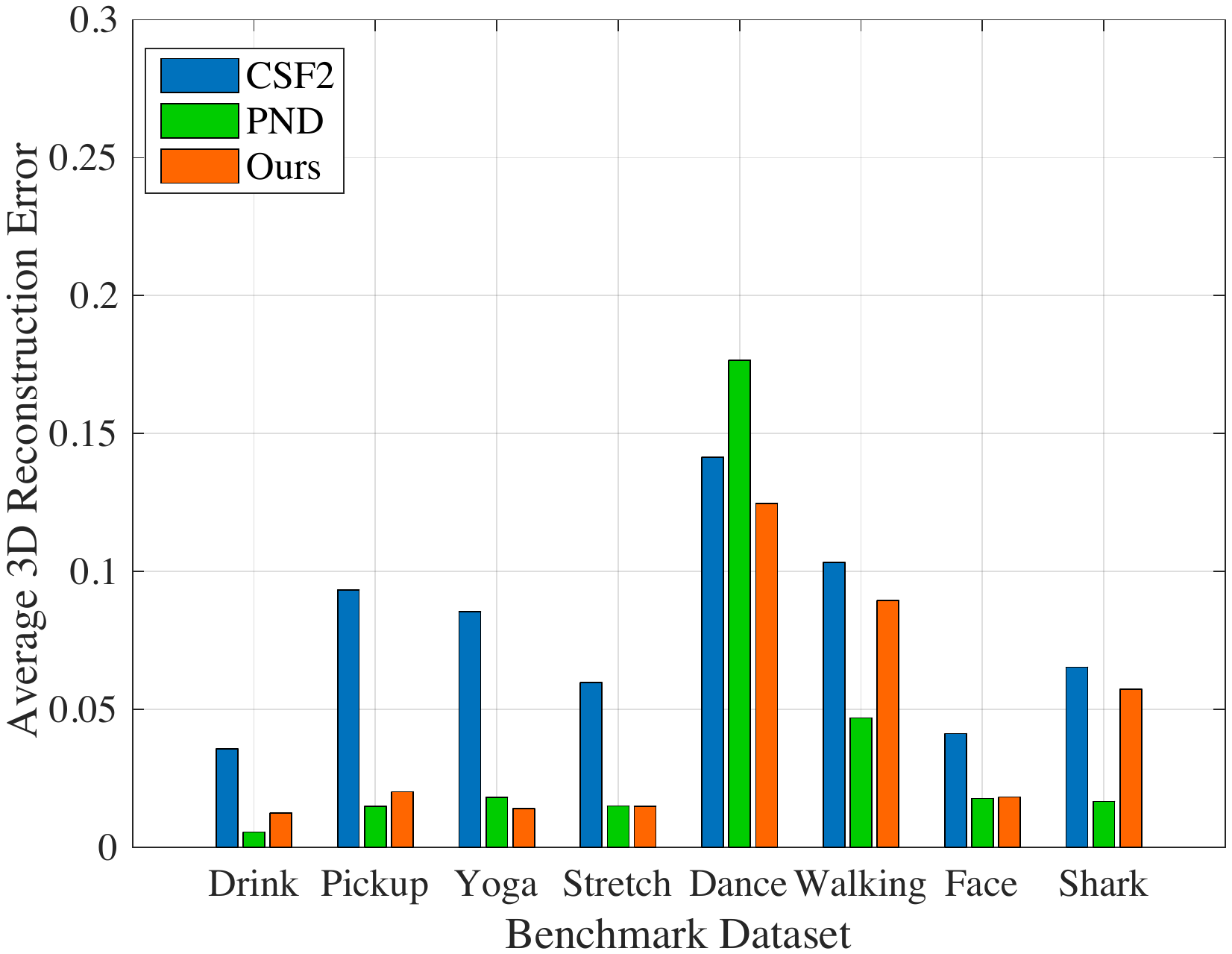}}
\caption{\small{(a) 3D reconstruction error comparison over noisy trajectories. (b) Comparison of our method performance with other competing methods with missing data in the measurement matrix. Note: BMM was not formulated for missing data case, therefore, its results are not present in the above figures.}}
\label{fig:evaluationNoiseandMD}
\end{figure}

\noindent
{\bf{3. Noisy data:}}\label{ss:noisy}
The feature tracks captured from a real-world motion capture system is noisy most of the time. Therefore, to test the reliability and robustness of our new approach, we performed experiments by re-synthesizing the trajectories added with Gaussian noise. We introduced the Gaussian noise with standard deviation set as $\sigma_\textrm{noise} = r*\textrm{max}\{|\m W|\}$, where $r$ is varied from 0.05-0.25 \cite{lee2013procrustean}. Figure (\ref{fig:n1}) shows the variation in the normalized average 3D error for the stretch sequence using the performance of different algorithm recorded over 20 times. The plot clearly shows the robustness of our algorithm in comparison to other methods in the presence of large noise ratio's. \\
\noindent
{\bf{4. Missing Data:}}\label{ss:missing}
In addition to the noisy data, the other problem with 3D reconstruction from a real video sequence is the missing trajectories over frames. We handle the missing trajectory quite robustly by incorporating a simple modification to the optimization proposed in Eq:(\ref{eq:10}). Let's assume $\tilde{\m W} \in \mathbb{R}^{2\m F \times \m P}$ is the incomplete measurement matrix and $\m M \in \{0, 1\}$ is the mask matrix which indicates the presence or absence of the tracks over frames. Given $\tilde{\m W}$, $\m M$, we first find a complete $\m W$ matrix using the following  optimization
\begin{equation}
\begin{aligned}
& \displaystyle \underset{\m W} {{\fontfamily{cmtt}\selectfont \text{minimize}}} ~\| \m M \odot(\tilde{\m W} - \m W)\|_{\m F}^{2}, ~{\fontfamily{cmtt}\selectfont \text{subject to:}} ~{\fontfamily{cmtt}\selectfont \text{rank}}(\m W) \leq 3\m K
\end{aligned}
\end{equation}
The above optimization is a well studied optimization form. To keep things simple, we used Cabral \etal work \cite{cabral2013unifying} to estimate $\m W$. The motive is to first solve for complete `$\m W$' to estimate camera motion using our rectified approach \S \ref{ss:rrot}, and then solve for shape using the following cost function:
\begin{equation}
\begin{aligned}
& \displaystyle \underset{\m S^{\sharp}, \m S} {{\fontfamily{cmtt}\selectfont \text{minimize}}} ~ \mu \|\m S^{\sharp}\|_{\Theta, *} + \frac{1}{2}\| \m M \odot(\tilde{\m W} - \m R \m S)\|_{\m F}^{2}\\
& \displaystyle {\fontfamily{cmtt}\selectfont \text{subject to:}} ~\m S^{\sharp} = g(\m S)
\end{aligned}
\end{equation}
% \begin{figure*}
% \centering
% \includegraphics[width=1\textwidth] {Figures/Result1Challenge.pdf}~~~
% \caption{\small{Reconstruction results of our method on the NRSfM challenge dataset \cite{jensen2018benchmark}. The results shown here are for the circular camera path. Ground-truth and reconstructed points are shown in filled and non-filled circles respectively.}}
% \label{fig:challengeResult1}
% \end{figure*}

Clearly, it's just a minor adjustment to the proposed method based on the kind of data available in different situations. To evaluate our performance, we randomly set 30\% of the data missing from the sequence same as Lee \etal work \cite{lee2013procrustean} for comparison. Figure (\ref{fig:m1}) shows the performance of our algorithm with missing data.
%Supplementary video is provided to show results on real world missing data cases.
%We provide a supplementary video to present the missing data results on some popular movie clips. We didn't show its results on this document to avoid copyright issues.

\noindent
{\bf{Discussion:}}
%In some applications, we have more prior knowledge about the shape in addition to its low-rank matrix assumption, for example: exact rank of the clean shape matrix. In such cases, one may choose to minimize partial sum minimization of singular values optimization \ie,
%\begin{equation}
%\begin{aligned}
%& \displaystyle \underset{\m S^{\sharp}, \m S} {{\fontfamily{cmtt}\selectfont\text{minimize}}} ~ \mu |{\fontfamily{cmtt}\selectfont \text{rank}}(\m S^{\sharp}) - \m T| + \frac{1}{2}\| \m W - \m R \m S\|_{\m F}^{2}\\
%& \displaystyle {\fontfamily{cmtt}\selectfont \text{subject to:}} ~\m S^{\sharp} = g(\m S)
%\end{aligned}
%\end{equation}
%where, $\m T$ is the target rank of the shape matrix. However, such an optimization needs an introduction to new operator known as PSVT \cite{oh2016partial} to optimize the problem. Nevertheless, PSVT can be regarded as special case of solving the weighted nuclear norm minimization \cite{chen2013reduced,gaiffas2011weighted}. Therefore, the point is, depending on the application, the proposed approach can be modified or changed, hence, its flexible.
\emph{Why not add the motion regularization $\|R_{t} - R_{t-1}\|_{F}$ in the final optimization and solve for both motion and shape?}
%It's definitely a valid argument. However, our attempt is to make ``prior-free approach'' \cite{dai2014simple} practically more suitable, therefore,  we followed the two staged approach \ie, first solve for motion and then solve for shape. Detailed analysis on how similar the smooth motion trajectories are: the one obtained from Intersection theorem and the other after adding such motion regularization. All these investigations of our algorithm are left for further extension.
It’s definitely a valid argument. Nevertheless, we wanted to show the competence
in a ``prior free'' way \cite{dai2014simple} which is to ``solve for motion first using Intersection theorem and then solve for low-rank shape'', therefore, we avoided to add it in the final optimization. We showed that smooth solution \cite{rabaud2008re} already exist within the solution to intersection theorem. Comprehensive analysis of our algorithm after adding motion regularization to solve the final optimization is left as an extension to the present idea.

%{\scriptsize{Kindly go through the supplementary material for ablation study, results on dense sequences and more algorithmic analysis.}}

\section{Conclusion}
\noindent
With weighted nuclear norm minimization of the shape matrix and an analytic solution to the rotation matrix based on the smoothness of the camera motion \cite{rabaud2008re}, we witnessed that the prior-free {\bf{idea}} performs almost as good as the best available algorithm's. Without exploiting the ``prior-free'' idea \cite{dai2014simple} \emph{fully} based on the well-known assumptions of smooth deformation of the non-rigid object and its low-rank shape, it may perform badly, which might be the reason that researchers have had poor results using it, even for the non-rigid objects that span a single linear subspace. Our work revealed the possibility of making ``prior-free'' \cite{dai2014simple} practically more accurate under the different conditions of measurement matrix with elementary modifications, and also conjecture some open problems. The accuracy of our algorithm on the benchmark datasets empirically validates that the ``prior-free'' theory is still a very powerful way to solve NRSfM and therefore, the \textbf{proposition} before the NRSfM researchers to consider is, it's not the failure of the \emph{concept} behind the prior-free idea for its inferior performance but, it's possibly due to our inability to correctly cater, and cleverly exploit the arc of information and perspectives provided by it to solve NRSfM.\\
%{\footnotesize{\textcolor{black}{Note: Movie clips in the supplementary material are used solely for research purpose, its content may subject to copyright}.}}

\noindent
\textbf{Acknowlegment}: The author research work is supported by Google and ETH Z\"urich Foundation project number 2019-HE-323 (2).

{\small
\bibliographystyle{ieee}
\bibliography{NRSFM_WACV20}
}

\end{document}